\documentclass[letterpaper,journal]{IEEEtran}
\usepackage{amsmath,amsfonts}
\usepackage{algpseudocodex}
\usepackage{algorithm}
\usepackage{amssymb}
\usepackage{tabularx}
\usepackage{color}
\usepackage{array}
\usepackage{makecell}
\usepackage[caption=false,font=normalsize,labelfont=rm,textfont=rm]{subfig}
\usepackage{textcomp}
\usepackage{stfloats}
\usepackage{url}
\usepackage{verbatim}
\usepackage{graphicx}
\usepackage{cite}
\usepackage{booktabs}
\usepackage{multirow}
\usepackage{orcidlink}
\hypersetup{hidelinks}

\begin{document}
\definecolor{mycolor}{RGB}{110,154,155} 
\title{Evaluating and Correcting Human Annotation Bias in Dynamic Micro-Expression Recognition}

\author{Feng Liu$^{\orcidlink{0000-0002-5289-5761}}$,~\IEEEmembership{Senior Member,~IEEE}, Bingyu Nan$^{\orcidlink{0009-0001-7235-2617}}$, Xuezhong Qian$^{\orcidlink{0009-0009-9850-4154}}$, and Xiaolan Fu$^{\orcidlink{0000-0002-6944-1037}}$,~\IEEEmembership{Member,~IEEE}
\thanks{This work was supported by National Key Research and Development Program of China (2024YFC3606801), by Shanghai Jiao Tong University 2030 Initiative and by Startup Fund for Young Faculty at SJTU (SFYF at SJTU, Grant No. 25X010506040). (\emph{Primary Corresponding author: Prof. Feng Liu, Corresponding author:Prof. Xiaolan Fu})}

\thanks{Prof.Feng Liu and Prof.Xiaolan Fu are with
School of Psychology, Shanghai Jiao Tong University, Shanghai 200030, China. (e-mail: liu.feng@sjtu.edu.cn;fuxiaolan@sjtu.edu.cn).
Bingyu Nan and Prof.Xuezhong Qian are with the
School of Artificial Intelligence and Computer Science, Jiangnan University, Wuxi 214122, China. (e-mail: 6233151001@stu.jiangnan.edu.cn;xzqian@jiangnan.edu.cn).
}}
\markboth{IEEE Transactions on Affective Computing, 2026, Vol. XX, No. XX}%
{Shell \MakeLowercase{\textit{Feng Liu et al.}}}


\maketitle

\begin{abstract}
 Existing manual labeling of micro-expressions is subject to errors in accuracy, especially in cross-cultural scenarios where deviation in labeling of key frames is more prominent. To address this issue, this paper presents a novel Global Anti-Monotonic Differential Selection Strategy (GAMDSS) architecture for enhancing the effectiveness of spatio-temporal modeling of micro-expressions through keyframe re-selection. Specifically, the method identifies Onset and Apex frames, which are characterized by significant micro-expression variation, from complete micro-expression action sequences via a dynamic frame reselection mechanism. It then uses these to determine Offset frames and construct a rich spatio-temporal dynamic representation. A two-branch structure with shared parameters is then used to efficiently extract spatio-temporal features. Extensive experiments are conducted on seven widely recognized micro-expression datasets. The results demonstrate that GAMDSS effectively reduces subjective errors caused by human factors in multicultural datasets such as SAMM and 4DME. Furthermore, quantitative analyses confirm that offset-frame annotations in multicultural datasets are more uncertain, providing theoretical justification for standardizing micro-expression annotations. These findings directly support our argument for reconsidering the validity and generalizability of dataset annotation paradigms. Notably, this design can be integrated into existing models without increasing the number of parameters, offering a new approach to enhancing micro-expression recognition performance. The source code is available on GitHub[https://github.com/Cross-Innovation-Lab/GAMDSS].
\end{abstract}

\begin{IEEEkeywords}
micro-expression, frame reselection, dynamic micro-expression recognition, differential selection strategy, computational perception
\end{IEEEkeywords}

\section{Introduction}
{F}{acial} expressions are considered to be one of the most fundamental and direct non-verbal signals through which humans express their inner emotions and psychological states\cite{CVPR2022zeng_fer}. They are widely regarded as a crucial window into an individual's true thoughts and underlying intentions\cite{TMM2023ben}. Facial expressions are generally categorized into two types: macro-expressions and micro-expressions. Macro-expressions are characterized by their higher intensity and longer duration, typically exceeding 0.5 seconds, and are readily discernible\cite{CVPR2023wang_DFER, MM2024li_DFER}. In contrast, micro-expressions are distinguished by their lower intensity, extremely brief duration, typically between 1/25 and 1/5 seconds, and regional specificity\cite{mm2022lu, TMM2023chen}. They constitute involuntary facial reactions resulting from emotional leakage when an individual attempts to suppress or conceal genuine emotions. It is precisely this authenticity, which is difficult to feign, that renders micro-expressions a key clue for revealing true affective states, thereby holding significant research value and application potential in fields such as clinical psychology, national security, and forensics\cite{mm2023guo, mm2023yang, aaai2023cmnet}.

\begin{figure}[!t]
    \centering
    \includegraphics[width=0.9\linewidth]{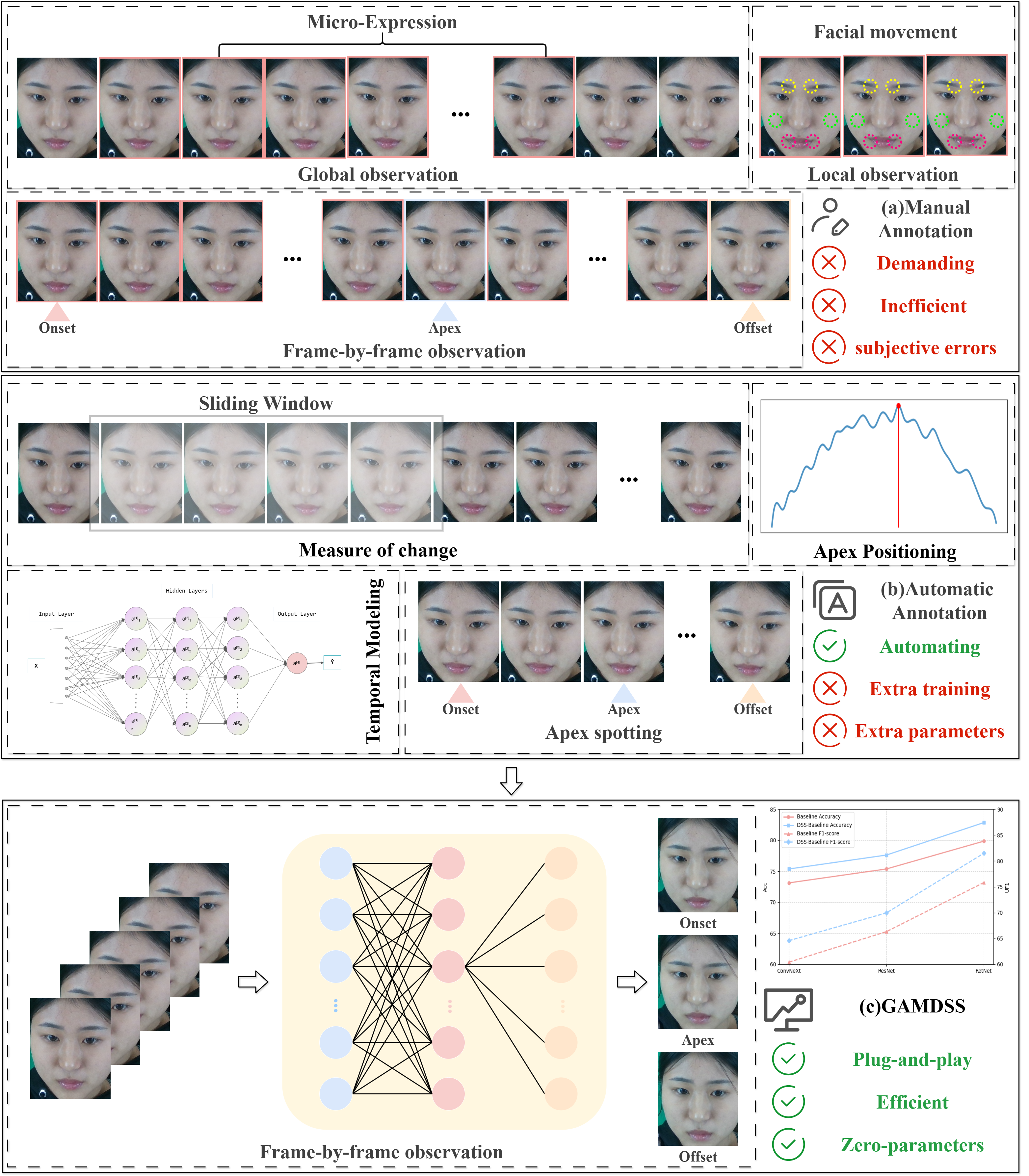}
\caption{This paper attempts to address the distortion of ground truth labeling due to human subjective errors in the annotation of micro-expression datasets. (a) Traditional manual annotation of key frames involves three steps: global observation, local observation, and frame-by-frame observation. However, this process requires high expertise from annotators, and the frame-by-frame observation stage may introduce subjective errors. (b) The automatic annotation method locates the Apex by calculating inter-frame changes through a sliding window, without requiring manual frame-by-frame comparison. However, this method typically relies on an extra training process and introduces extra model parameters. (c) The proposed GAMDSS method can re-select key frames based on manual annotations, effectively avoiding human subjective errors. Importantly,  it is plug-and-play and does not introduce additional parameters.}
\label{fig:concept}
\end{figure}

Micro-expression analysis techniques include micro-expression spotting and micro-expression recognition. Micro-expression spotting involves detecting segments that may contain micro-expressions in a video and, upon confirming their presence, annotating the onset, apex, and offset frames\cite{tipwang2021, mm2024Yu}. Micro-expression recognition involves the classification of the emotions depicted in the annotated micro-expression segments\cite{TOMM2023gong}.

\begin{figure*}[!htbp]
    \centering
    \includegraphics[width=0.85\linewidth]{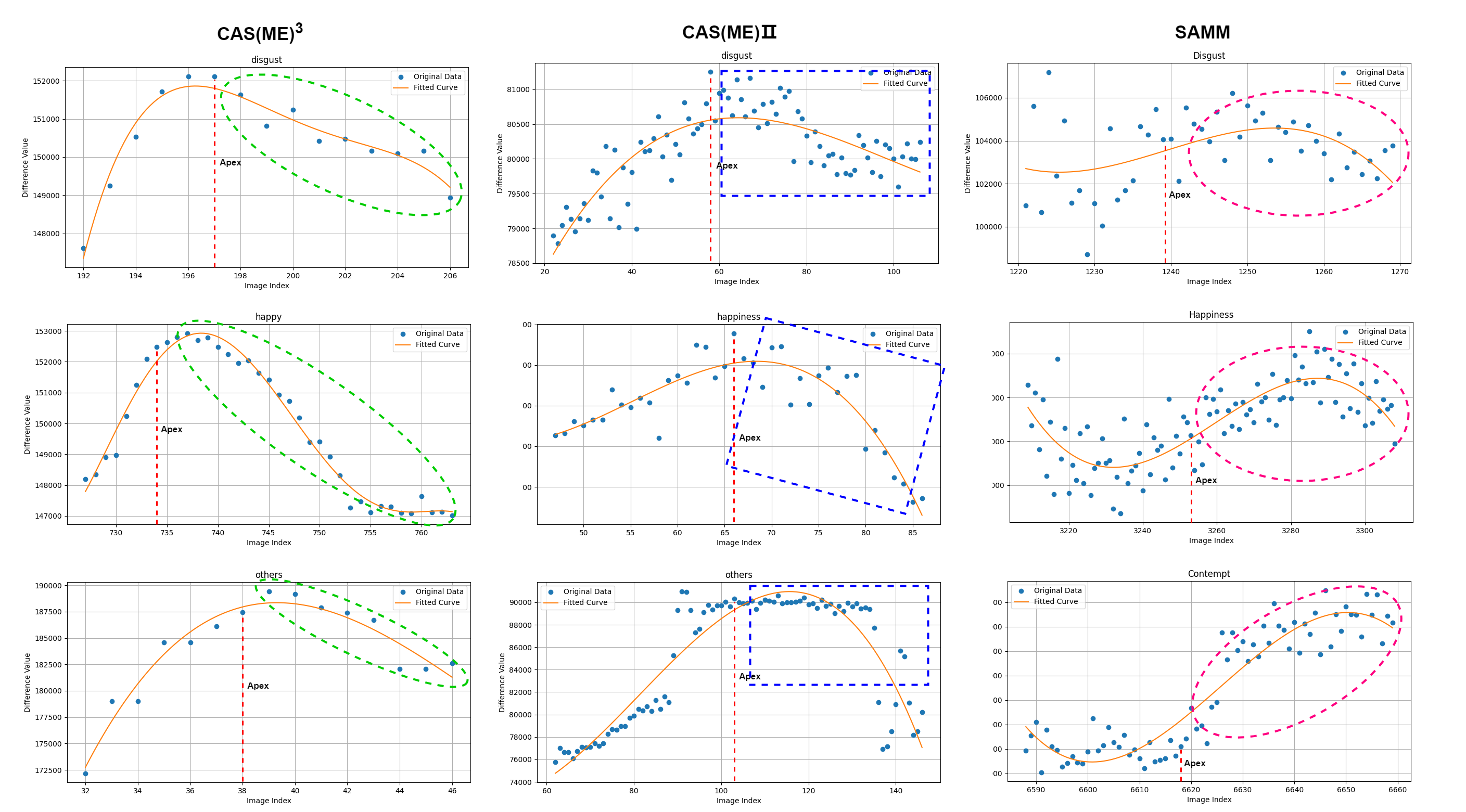}
\caption{
Under the three-classification conditions, the individual sample difference curves obtained after calculating differences across three datasets are shown in the figure. Specifically, the L2 norm of pixel values between frames is computed frame-by-frame to quantify motion intensity, with its apex serving as a key objective metric for assessing changes in expression intensity. The original manually annotated Apex frames in the dataset are marked with red dashed lines, while the maximum difference values derived from the difference calculations are highlighted with dashed boxes.
}
\label{fig:curve}
\end{figure*}

Early micro-expression recognition research primarily focused on analyzing complete video sequences, which resulted in time-consuming analysis processes and data redundancy. Subsequent studies analyzed the micro-expression recognition process from a cognitive psychology perspective\cite{ekman1993}. They found that when the intensity of a micro-expression reaches its Apex, the information it conveys reflects the current emotional state. Subsequent studies experimentally validated this conclusion and proposed using the Apex frame rather than the entire sequence as the foundation for micro-expression recognition tasks\cite{liong2017}. However, another issue is that micro-expressions occur very briefly, typically between 1/25 and 1/5 of a second, which makes manual Apex frame labeling difficult\cite{TIPli}. To improve human detection of micro-expressions, psychologist Ekman and his team developed a special micro-expression training tool (METT). However, it is worth noting that even after systematic training with METT, manual micro-expression detection accuracy remains low, typically not exceeding 50\%\cite{2024review}.

Specifically, due to variations in annotation scenarios, significant discrepancies exist between the manually annotated Apex frames and ground-truth apex frames. For the sake of brevity, only the difference curves from three representative datasets are displayed, along with samples from the three categories: negative, positive and surprise. Results for negative and positive categories in CASME \uppercase\expandafter{\romannumeral2} and CAS(ME)$^3$ indicate that the action intensity generally follows a smooth decay curve after the annotated Apex frame. Although there are differences between the annotated Apex frames and the ground-truth Apex frames, datasets collected in a single cultural context reduce the discrepancies between the two. Conversely, datasets collected in multicultural contexts, such as SAMM, show greater frame differences. As shown in Fig. \ref{fig:curve}, in both negative and positive results, significant fluctuations in action intensity occur after the annotated Apex frame, rather than a smooth decline in action intensity. This indicates that datasets more commonly found in real-world environments will exhibit greater frame differences, leading to increased manual annotation noise introduced by labels during the learning process. This issue has strict limitations, and if resolved, it would allow for optimization and enhancement of bottlenecks based on the annotation level of the dataset.

To better understand the source of annotation noise, we examine the manual annotation process. Manual annotation of micro-expressions is broadly divided into three stages: global observation, local observation, and frame-by-frame analysis. During the global observation, annotators identify video segments containing micro-expressions within raw long-duration videos. In the local observation, annotators precisely annotate the specific locations where actions occur. Finally, during frame-by-frame analysis, annotators focus on specific facial regions and annotate the onset, apex, and offset frames by comparing action changes frame by frame. This stage is particularly susceptible to subjective influences. Research has indicated that mastering this manual annotation process requires over 100 hours of systematic training.

To address the above problems, we propose a novel global anti-monotonic differential selection strategy architecture, GAMDSS. Specifically, the method is able to capture Onset frames and Apex frames characterized by maximum micro-expression changes from complete dynamic sequences, and determine Offset frames based on them. This process effectively reduces the subjective errors of manual labeling and captures more fine-grained spatio-temporal action information. Previous methods usually take only the Onset and Apex keyframes as inputs, ignoring the dynamic change information from the Apex to the Offset phase, thereby limiting the understanding of the complete change process of the micro-expression actions. To address this limitation and further explore the reasons for the abnormal fluctuations in the action intensity curves following the manually annotated Apex frames observed in the multicultural dataset, we designed two spatio-temporal units with shared parameters to extract complete spatio-temporal dynamic features in order to achieve comprehensive modeling of spatio-temporal relationships.

As demonstrated above, the particular core contributions of the present paper may be enumerated as follows.
\begin{itemize}

\item To the best of our knowledge, this is the first micro-expression oriented study to address the issue of distortion in ground truth labeling caused by human subjectivity. Rather than improving the model itself directly, this study explores performance enhancement paradigms that can be seamlessly integrated into existing systems. This is achieved by thoroughly analyzing the boundaries of ground truth labels and designing search strategies accordingly.

\item We propose the Global Anti-Monotonic Difference Selection Strategy (GAMDSS) architecture. This architecture dynamically captures the three most discriminative key frames from complete micro-expression sequences, subsequently constructs comprehensive spatio-temporal dynamic features, and extracts them through dual spatio-temporal units with shared parameters. This integrated spatio-temporal relationship modeling process enables adaptive adjustment to original manual annotation biases.

\item Results demonstrate that GAMDSS effectively reduces subjective annotation errors across multicultural datasets like SAMM and 4DME. Adaptively calibrated outputs achieve state-of-the-art (SOTA) performance, significantly outperforming other non-pre-trained models. Quantitative analysis further confirms: for single culture micro-expression datasets like CASME \uppercase\expandafter{\romannumeral2} and CAS(ME)$^3$, onset and apex frames suffice to capture most micro-expression change features. However, this assumption no longer holds for cross-cultural micro-expression datasets like SAMM and 4DME.

\item This research introduces a new fundamental paradigm, which can be applied to any time series labeling alignment problem. This has great potential to improve the accuracy of existing deep learning models that rely on time series labels, without affecting the original model architecture.

\end{itemize}

\section{Related Works}
\label{sec:Related Works}
\subsection{Micro-expression Recognition}

Traditional micro-expression recognition technology primarily relies on manually extracted features of micro-expressions for identification. For example, some studies have proposed combining Local Binary Pattern histograms from Three Orthogonal Planes (LBP-TOP) with traditional classifiers to address the micro-expression recognition task\cite{li2013lbp}. Building on this foundation, subsequent research has focused on expanding and optimizing the original LBP-TOP operator, with the core objective of reducing the redundancy of the LBP-TOP operator to enhance recognition accuracy\cite{wang2015}. In this process, new operators based on LBP-TOP, such as SCCLQP and HIGO-TOP, have been proposed\cite{SCCLQP, HIGO-TOP}. Additionally, new features such as MDMO have been introduced to enhance the competitiveness of the model\cite{MDMO}.

With the significant achievements of deep learning in the field of computer vision, its powerful performance in facial recognition has sparked a wave of research into its application in micro-expression recognition tasks. A typical example is the work of \cite{patel2015}, which has adapted the convolutional neural network previously used for face recognition to micro-expression recognition. However, due to the limited size of micro-expression datasets, the model did not achieve the expected results. To optimize the performance of micro-expression recognition models, several innovative studies have been applied to the micro-expression recognition field in recent years. One study extended the temporal-spatial connectivity of CNN from both appearance and geometric perspectives to model the temporal-spatial deformation of micro-expression sequences \cite{TMM2020xia}. Another study learned feature mappings to map the features of micro-expression datasets into the feature space of macro-expression datasets, thereby leveraging the abundant annotated information in macro-expression datasets to enhance micro-expression recognition performance \cite{TMM2023ben}. Additionally, another study calculated four optical flow features between the Onset frame and the Apex frame in a video, divided the image into multiple blocks, and performed convolution on these blocks in a deep learning model to capture detailed micro-expression features\cite{TMM2023chen}. Although the aforementioned studies validated the temporal relationship of micro-expression recognition from the Onset frame to the Apex frame, they did not consider the complete temporal-spatial relationship or manual annotation errors. 

\subsection{Micro-expression Spotting}

Initial MES technology has primarily relied on manually designed features combined with sliding window or threshold segmentation strategies. For example, some studies have used the chi-square distance of local binary patterns (LBP) in a fixed-length scanning window to spot micro-expression segments\cite{Moilanen2014}. Other studies have utilized optical flow features, calculating the maximum difference magnitude along the principal direction to detect micro-expressions\cite{he2020}. Additionally, some research has employed changes in the Euclidean distance between facial landmarks in three facial regions to detect micro-expression segments\cite{beh2019}. However, these traditional methods have lacked robustness in complex scenarios and have been particularly susceptible to interference from head movements. As a result, more researchers have increasingly turned to deep learning techniques to overcome the limitations of traditional methods. Early attempts have included directly applying or fine-tuning existing models, such as having used CNN to distinguish neutral frames from Apex frames, then having merged adjacent detection samples through feature engineering methods\cite{zhang2018}. Alternatively, ensemble learning combined with multi-scale sliding windows has been used to detect micro-expression segments\cite{liu2024}. To more effectively model temporal patterns, some studies have extracted directional optical flow histograms (HOOF) to encode motion changes in selected facial regions, then have used RNN to identify segments potentially containing micro-expressions\cite{Verburg2019}. Building on this, other researchers have further utilized graph convolutional networks (GCN) to embed action unit (AU) information into the extracted optical flow, thereby having learned complex spatial relationships\cite{yin2023}. Although these methods have made some progress, due to the limited scale of micro-expression datasets and the inherent subtlety and transience of micro-expressions, their detection performance still has lagged significantly behind existing manually annotated levels.

\subsection{Frame Selection Methods}

In the domain of micro-expression recognition, Apex frames have been shown to provide the primary emotional information, thus emphasizing the importance of accurate identification of start and apex frames in capturing key moments of expression changes. Despite the existence of manually annotated key frames in existing datasets, such as CASME \uppercase\expandafter{\romannumeral2}, SAMM and CAS(ME)$^3$, studies have demonstrated that these annotations may be subject to subjective errors\cite{2024review}. To improve recognition accuracy, some studies have used LBP to extract texture information from regions of interest (ROI) on the face and have identified Apex frames by measuring frame-to-frame differences\cite{yan2015}. Other studies have utilized integrated optical flow vectors calculated in small local spatial regions to identify initial and Apex frames in videos\cite{patel2015}. Furthermore, some studies have employed a binary search strategy to process several subregions of the facial ROI using LBP and optical strain, thus locating key frames\cite{li2016hoof}. Meanwhile, Some studies have defined 26 facial regions, have extracted 3D HOG features for each region, and have used the chi-square distance to detect key frames\cite{Davison3dhog}. Unlike the aforementioned methods, some studies have improved the performance of Apex frame localization by using a 3D fast Fourier transform (3D-FFT) in the frequency domain to detect Apex frames\cite{li2021}. Although these methods have been effective for automatically localizing key frames, the accuracy of identifying subtle and rapid micro-expression changes has still needed improvement. Recent studies have also attempted to avoid reliance on apex frame annotations and enhance robustness to noise through self-supervised learning and local deformation modeling. However, their performance has remained limited by sensitivity to hyperparameters and alignment capabilities in complex scenes\cite{zhang2025cvpr}.

To address this issue, we propose a novel frame selection strategy (GAMDSS) that constructs complete temporal dynamic features by capturing the three frames with the most significant micro-expression changes in the entire dynamic video stream. Unlike the MES method, our method re-annotates based on manual annotation. The core purpose of this strategy is to effectively mitigate subjective errors that arise during manual annotation, thereby providing the model with more refined spatio-temporal action information.

\begin{figure*}[!t]
    \centering
    \includegraphics[width=0.85\linewidth]{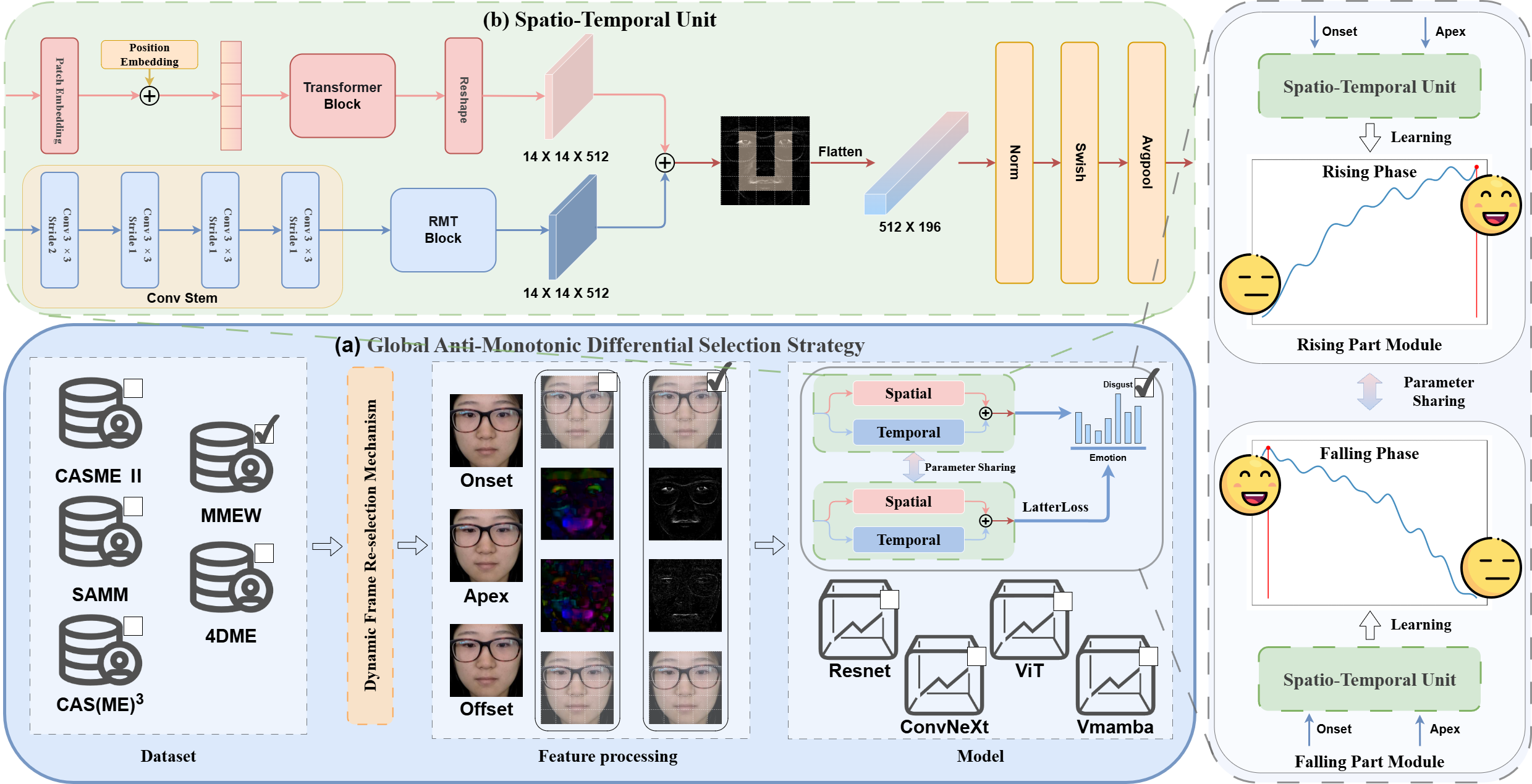}
\caption{An overview of the proposed GAMDSS architecture is provided below. (a) The GAMDSS pipeline consists of the following steps: First, Dynamic Frame Reselection Mechanism reselects the three frames with the richest action changes based on different datasets. Second, a backbone model and feature processing method are selected. Next, spatio-temporal features are extracted at different stages using spatio-temporal units with two shared parameters. Where the temporal stream integrates the RMT module, which efficiently models long-term temporal dependencies through a retention mechanism based on Manhattan distance decay. Finally, the spatio-temporal features are integrated, and an auxiliary loss function is introduced to inject additional knowledge, thereby enabling the modeling of the complete evolution process of micro-expressions. (b) The designed method for extracting spatio-temporal features and their fusion approach, where Swish activation layers are employed to enhance feature nonlinearity and improve optimization stability. }
\label{fig:architecture}
\end{figure*}

\section{Methodology}
\label{sec:Methodology}

\subsection{Overview}

In this section, a thorough description of the proposed GAMDSS is provided. As demonstrated in Fig. \ref{fig:architecture}, the core design of this architecture is predicated on a dynamic frame re-selection mechanism. This mechanism performed local searches in proximity to the original manually annotated key frames through differential calculations to identify the three key frames with the most significant action changes. Subsequently, two spatio-temporal units with shared parameters were employed to extract spatio-temporal features from different stages. Finally, spatio-temporal features were integrated, and an auxiliary loss function was introduced to inject additional knowledge, enabling the modeling of the complete evolution process of micro-expressions.

\subsection{Dynamic Frame Re-selection Mechanism}
\begin{figure}[!t]
    \centering
    \includegraphics[width=0.85\linewidth]{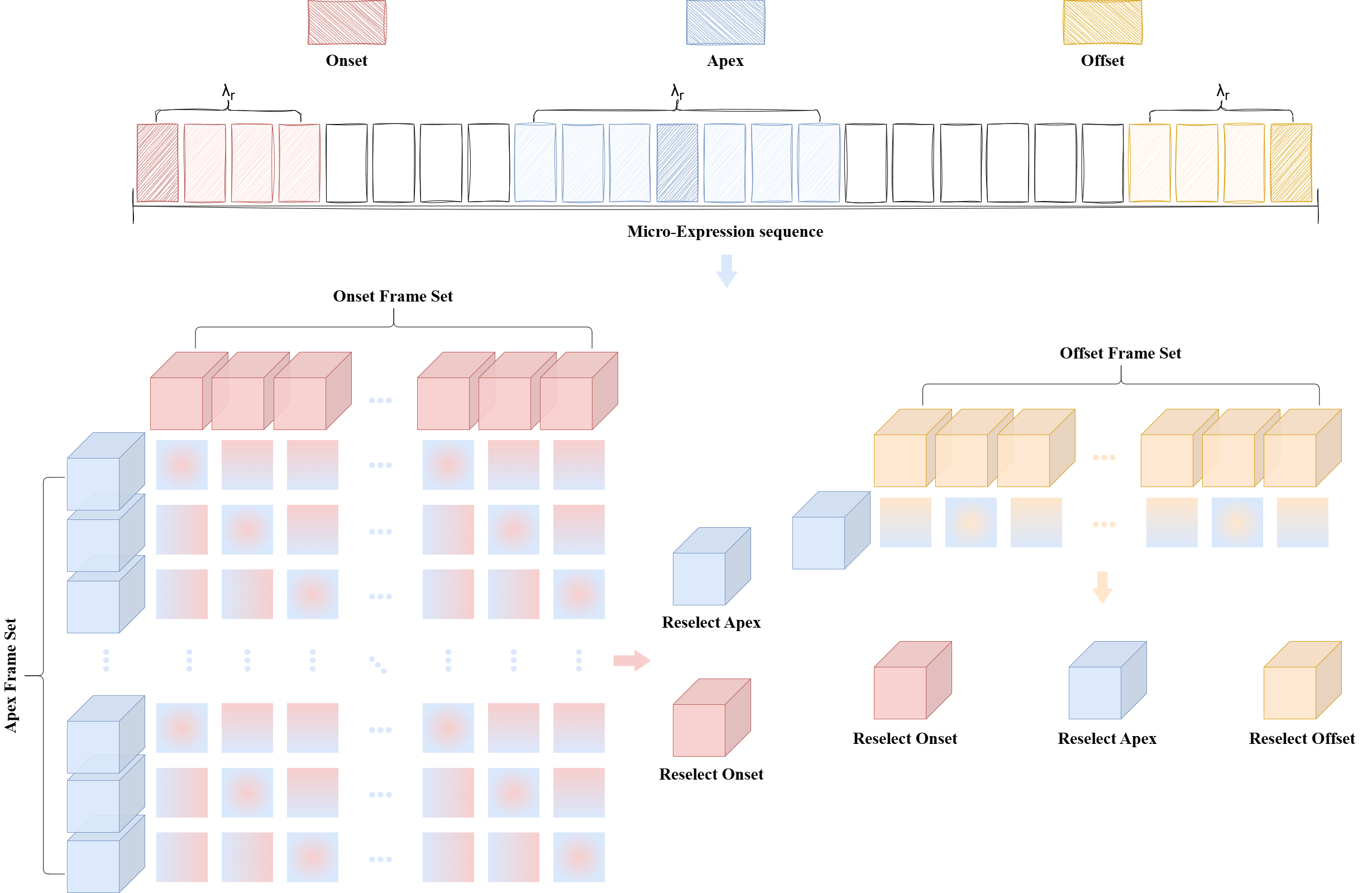}
\caption{Conceptual design of GAMDSS. First, a set of relevant frames is determined based on manually annotated key frames. Second, frame pairs with the greatest action changes are reselected from the frame set through difference calculation, and these are used as the reselected onset frames and apex frames. Finally, the reselected offset is determined based on the reselected apex frames.}
\label{fig:GAMDSS}
\end{figure}

In the given time series, the objective was to identify the three frames that exhibit the most significant changes and designate them as the 'Onset', 'Apex' and 'Offset', respectively. However, directly filtering these three frames from the entire time series imposed a substantial computational burden. To address this challenge, a two-step approach was adopted. Initially, a local search region is defined around the manually annotated onset frame and apex frame. By calculating the frame differences between all frame pairs within this region, the most significant difference pair is selected as the reselected onset frame and apex frame. Subsequently, using the newly determined apex frame as a reference, the same difference method is applied to subsequent frames in the sequence to locate offset frames that best characterize expression decay features, as illustrated in Figure \ref{fig:GAMDSS}. The transition from a state of calm to Apex intensity was defined as 'rise', and the transition from Apex intensity to calm again was defined as 'fall'. In exceptional cases, the start frame could coincide with an Apex frame or an offset frame. To quantify this variation, it is assumed that the start frame is indexed by $Onset$ and the Apex frame is indexed by $Apex$. Subsequently, a selection range $R_{rise}$ was introduced to select a specific number of frames around the start and Apex frames:
\begin{equation}
    R_{rise} = (Apex-Onset)\lambda_{rise}
\end{equation}

Where $\lambda_{rise}$ is the scaling factor of the selection range, which determines the size of the selection range.

Once the selection range was determined, the frame difference was calculated. 

Specifically, we construct the set \( P \) of all candidate frame pairs within the local search range. Its total number of pairs, denoted by \(L = (R_{\text{rise}} + 1)(2R_{\text{rise}} + 1) \), is defined as follows:
\begin{equation}
    P = \{p_i\}_{i=1}^L = \{ (f_o, f_a) \}_{i=1}^L
\end{equation}

Where P is the set of frame pairs $p_i$, \text{$i \in [1,L]$}, and ($f_o$, $f_a$) denotes the selected Onset frames and Apex frames. \text{$o \in [onset_t,onset_t+R_{rise}]$}, and \text{$a \in [apex_t-R_{rise},apex_t+R_{rise}]$}.

Based on the frame pairs, we calculated the frame differences as follows:
\begin{equation}
\Delta f = \| f_a - f_o \|_2
\end{equation}

Where \text{$\Delta f$} denotes the frame difference, from which we subsequently select the largest frame differences and obtain their frame indexes as Onset frames and Apex frames as follows:
\begin{equation}
f(Onset, Apex) = max( \{p_i\}_{i=1}^L)
\end{equation}

Ultimately, the two frames showing the most significant changes were identified as onset and Apex frames in a given time series. The offset frames were then filtered using the filtered Apex frames obtained following the same method. The process is outlined in Algorithm \ref{algo:GAMDSS}.

\begin{algorithm}[ht!]
    \caption{Pseudo code for Dynamic frame reselection mechanism.}
    \begin{algorithmic}[1]
    \Require $f: micro-expression\,frame\,sequences;$
    \Require $onset: onset\, index;\, apex: apex\, index;\,offset: offset\, index$
    \Require $\lambda_{rise},\lambda_{fall}: scaling\,factor$
    \Ensure $index_{onset};\, index_{apex};\,index_{offset}$
    
    \State $R_{rise} \gets (apex - onset) \times \lambda_R$
    \State $R_{fall} \gets (offset - apex) \times \lambda_R$
    \For{$i \gets 1$ to $R_{rise}+1$}
        \State $diff_0 \gets \| f_a - f_{o+i} \|_2$
        \State $diff\_dict[(a,o+i)] \gets diff_0$
        \For{$j \gets 1$ to $R_{rise}+1$}
            \State $diff_1 \gets \| f_{a-j} - f_{o+i} \|_2$
            \State $diff\_dict[(a-j,o+i)] \gets diff_1$
        \EndFor
    \EndFor
    \For{$n \gets 1$ to $R_{rise}+1$}
        \State $diff_2 \gets \| f_{a-n} - f_{o} \|_2$
        \State $diff\_dict[(a-n,o)] \gets diff_2$
    \EndFor
    \State $onset, apex \gets \max(diff\_dict, key=diff\_dict.get)$
    \For{$m \gets 1$ to $R_{fall}+1$}
        \State $diff_3 \gets \| f_a - f_{offset+i} \|_2$
        \State $after\_diff\_dict[(a,{offset}+i)] \gets diff_3$
    \EndFor
    \State $ \_, offset \gets \max(after\_diff\_dict, key=after\_diff\_dict.get)$
    \State \Return $onset, apex, offset$
    \end{algorithmic}
    \label{algo:GAMDSS}
    
\end{algorithm}

\subsection{Spatio-Temporal Unit}
\textbf{Temporal Stream.} Retnet has been demonstrated to be an efficient architectural solution for language modeling\cite{retnet}. This work has proposed a retention mechanism for sequence modeling that has introduced temporal decay into sequence processing, thus facilitating the understanding of temporal relationships in the model. The retention mechanism has been retained as follows:
\begin{equation}
o_n = \sum_{m=1}^{n} \gamma^{n-m} (Q_n e^{i n \theta})(K_m e^{i m \theta})^\dagger v_m
\end{equation}

Where \(o_n\) is the output at time step \(n\). \(Q_n\), \(K_m\), and \(v_m\) represent the query, key, and value vectors, respectively. \(\gamma^{n-m}\) is a weighting factor that decays with increasing historical distance. Meanwhile, \(e^{i n \theta}\) and \(e^{i m \theta}\) represent the rotation position encodings injected into the key and query, respectively.

Since our core contribution is the proposed dynamic frame reselection mechanism rather than introducing a new micro-expression recognition network, using an established backbone network allows us to focus on the core issue without introducing unnecessary complexity.

\textbf{Spatial Stream.} In micro-expression recognition, temporal features have been shown to provide models with rich information about facial muscle movements. However, relying solely on temporal information made it difficult for models to understand the correspondence between the location of facial muscle movements and specific facial regions. To introduce this correspondence, we designed a spatial branch specifically to extract positional information. When the face is in a resting state, facial muscles do not undergo any deformation, providing stable facial position information. Therefore, we used the onset frame or offset frame as input features. Inspired by the ViT method, we divided the input features into 7 $\times$ 7 regions and added a learnable positional feature to each region, as illustrated in the following:

\begin{equation}
y_{b, d, h', w'} = \sum_{i=1}^{C} \sum_{j=0}^{P-1} \sum_{k=0}^{P-1} K_{d, i, j, k} \, x_{b, i, h' \cdot P + j, w' \cdot P + k} + b_{d}
\end{equation}

Where x is the input feature, P is the edge length of the patch, h' and w' are the indexes of the output feature graph indicating the position of the patch.

\begin{equation}
y_{b, n, d} = y_{b, n, d} + \text{pos\_embed}_{1, n, d}
\end{equation}

Where b, n and d denote the batch size, the number of patches and embedding dimension, respectively, pos\_embed denotes a set of learnable position tensors.

Subsequently, the processed features were fed into the Transformer block to obtain spatial features. It was worth noting that we used only two layers of Transformer block to avoid extracting the identity information that is not related to the micro-expression motion. It has been shown that the LayerNorm layer can introduces strong nonlinearities and such nonlinearities enhance the model's representational capabilities\cite{Zhu2025DyT}. In addition, this compression behavior reflects the saturation property of biological neurons for large inputs, a phenomenon that observed about a century ago. To introduce this strong nonlinearity, we added a LayerNorm layer before deforming the obtained tensor to a dimension size of 14 $\times$ 14 $\times$ 512 to enhance the characterization of spatial sub-branches.

Then, the tensors obtained from the spatial branch and the temporal branch were summed at the element level, which can be expressed as follows:

\begin{equation}
\mathcal{T_\text{s-t}} = \mathcal{T}_{\text{s}} + \mathcal{T}_{\text{t}}
\end{equation}

Where $\mathcal{T_\text{s}}$ is the spatial feature extracted from the spatial branch, $\mathcal{T_\text{t}}$ is the temporal feature extracted from the temporal branch, and $\mathcal{T_\text{s-t}}$ is the fused spatio-temporal feature.

\subsection{Global Anti-monotonic Differential Selection Strategy}
The Onset frame is defined as the initial frame at the commencement of the ME, whilst the Apex frame signifies the point at which the intensity of the facial expression attains its zenith. The disparity between these two pivotal frames encapsulated the action information, whereby the expression intensity progressively attained its maximum value. Subsequently, the discrepancy between the initial frame and the Apex frame was calculated based on GAMDSS, thereby yielding the difference frame as follows:
\begin{equation}
F_{rise} = F_{apex} - F_{onset}
\end{equation}

In a similar manner, the difference between the Apex frame and the Offset frame signifies the action change that the facial expression intensity apex and then gradually tends to calm down. Therefore, the difference between the Apex frame and the Offset frame obtained from the GAMDSS was calculated as follows:
\begin{equation}
F_{fall} = F_{apex} - F_{offset}
\end{equation}

Where $F_{rise}$ denote the difference frame in the Apex phase of the expression, $F_{fall}$ denote the difference frame in the calm phase of the expression, and $F_{onset}$, $F_{apex}$, and $F_{offset}$ denote the Onset, Apex, and Offset frames, respectively.

Subsequently, the two difference frames, $F_{rise}$ and $F_{fall}$, are computed by a Spatio-Temporal Unit (S-T Unit) with shared parameters to obtain spatio-temporal feature information, respectively. It is noteworthy that the spatio-temporal unit employed for the two different stages shares parameters to address the issue of limited data in the micro-expression recognition task. The feature extraction process is represented as follows:

\begin{equation}
\phi_{rise}, \phi_{fall} = \text{S-T Unit}\left( F_{rise}, F_{fall}; \theta \right)
\end{equation}

Where \(F_{rise}\) and \(F_{fall}\) represent the difference frames for the rise and fall phases, respectively, serving as inputs to the spatio-temporal unit; \(\phi_{rise}\) and \(\phi_{fall}\) denote the corresponding spatio-temporal features extracted by the unit for the two phases; \(\theta\) denotes the learnable shared parameters within this spatio-temporal unit.

Based on the extracted spatio-temporal features, we integrate the loss function of the falling phase into the model through knowledge injection to enhance its ability to recognize fine-grained spatio-temporal associations across the entire micro-expression sequence. This integration aims to promote a deeper understanding of the dynamic process: expression intensity gradually increases from a calm state, reaches an apex, and then begins to decrease from the apex. The total loss function during training is as follows:

\begin{equation}
L = -\sum_{i=1}^{C} t_i \cdot \log(p_{\text{rise},i}) - \sum_{i=1}^{C} t_i \cdot \log(p_{\text{fall},i})
\end{equation}

Where $C$ denotes the total number of categories, $t_i$ is the true label for the $i$th category, and $p_{\text{rise},i}$ and $p_{\text{fall},i}$ represent the model's predicted probabilities for the $i$th category during the rising and falling phases, respectively.

\begin{table*}[!t]
\caption{Statistical information on the mainstream micro-expression dataset.}
\centering
\begin{tabularx}{\textwidth}{@{}c|cccccccc}
\toprule
Dataset & Year & Subs & Gender (M/F) & Age & Samples & FPS & Emotions & Cultural composition \\
\midrule
CASME\cite{casme} & 2013 & 35 & 22/13 & Mean=22.03 & 195 & 60 & \makecell[c]{contempt disgust fear happiness \\ repression sadness surprise tense} & \makecell[c]{As: 35}\\
CASME \uppercase\expandafter{\romannumeral2}\cite{casme2} & 2014 & 35 & 22/13 & Mean=22.03 & 247 & 200 & \makecell[c]{disgust happiness repression surprise \\others} & \makecell[c]{As: 35}\\
SAMM \cite{samm} & 2016 & 32 & 16/16 & \makecell[c]{Mean=33.24\\(Range:19-57) } & 159 & 200 & \makecell[c]{anger contempt disgust fear \\happiness sadness surprise other} & \makecell[c]{E: 18 As: 8 ME: 2 \\Af: 3 Mix: 1} \\
CAS(ME)$^2$ \cite{cas(me)2} & 2017 & 22 & 9/13 & \makecell[c]{Mean=22.59\\(Range:19-26)} & 57 & 30 & \makecell[c]{anger disgust happiness} & \makecell[c]{As: 22}\\
MMEW \cite{mmew} & 2021 & 36 & / & Mean=22.35 & 300 & 90 & \makecell[c]{anger disgust fear happiness \\sadness surprise others} & \makecell[c]{As: 36}\\
4DME \cite{4dme} & 2022 & 56 & 38/27 & \makecell[c]{Mean=27.80\\(Range:22-57)} & 267 & 30 & \makecell[c]{negative positive surprise repression \\others} & \makecell[c]{As: 37  E: 28} \\
CAS(ME)$^3$ \cite{casme3} & 2022 & 247 & 112/135 & Mean=22.74 & 1109 & 30 & \makecell[c]{anger disgust fear happiness \\sadness surprise others} & \makecell[c]{As: 247}\\
\midrule
\multicolumn{9}{l}{E stands for Europeans, As stands for Asians, ME stands for Middle Easterners, Af stands for Africans, and Mix stands for mixed race.}
\end{tabularx}
\label{tab:dataset}
\end{table*}

\section{Experiments}
\label{sec:Experiments}
\subsection{Datasets and Evaluation Protocols}
To validate the effectiveness of GAMDSS, extensive experiments are conducted on the following representative micro-expression datasets.

\textbf{CASME}\cite{casme} contains 195 ME samples from 19 participants, with a frame rate set to 60 FPS. It is widely recognized as the original micro-expression dataset.

\textbf{CASME \uppercase\expandafter{\romannumeral2}}\cite{casme2} is an improved version of the CASME dataset, features higher frame rates and facial region resolution, enabling more precise capture of micro-expression changes.

\textbf{SAMM}\cite{samm} includes 159 ME samples from 32 participants, with a frame rate set to 200 FPS. Unlike other datasets, it includes participants from diverse cultural backgrounds, supporting cross-cultural micro-expression research.

\textbf{CAS(ME)$^2$}\cite{cas(me)2} includes 57 ME samples from 22 participants, with a frame rate of 30 FPS. There are three emotional categories.

\textbf{MMEW}\cite{mmew} consists of 300 ME and 900 MaE samples from 36 participants. The frame rate is set to 90 FPS. Each sample is labeled with seven emotional tags.

\textbf{4DME}\cite{4dme} innovates in data collection, containing multimodal video data from 56 participants, including 4D facial data, traditional 2D facial grayscale data, RGB video data, and depth video data. Additionally, it is important to note that its participants also come from diverse cultural backgrounds.

\textbf{CAS(ME)$^3$}\cite{casme3} dataset employs second- and third-generation elicitation paradigms to obtain ME samples with high ecological validity. It includes 1,109 ME samples from 247 participants, with a frame rate set at 30 FPS.


\begin{table*}[!t]
\caption{Performance comparison on CASME \uppercase\expandafter{\romannumeral2} and SAMM, ACC. stands for accuracy, and UF1 stands for unweighted F1 score. The best results are highlighted in bold, and the second-best results are underlined.}
\centering
\begin{tabular}{@{}c|cc|cc|cc|cc|c|c|c}
\toprule
\multirow{3}{*}{Model} & \multicolumn{4}{c|}{CASME \uppercase\expandafter{\romannumeral2}} & \multicolumn{4}{c|}{SAMM} & \multirow{2}{*}{\#param.} & \multirow{2}{*}{FLOPs} \\
& \multicolumn{2}{c|}{5-class (\% ↑)} & \multicolumn{2}{c}{3-class (\% ↑)} & \multicolumn{2}{|c|}{5-class (\% ↑)} & \multicolumn{2}{c|}{3-class (\% ↑)} \\
& ACC. & UF1 & ACC. & UF1 & ACC. & UF1 & ACC. & UF1 & & \\
\midrule
Swin-T\cite{swin} '21 & 67.21 & 62.47 & 84.42 & 79.87 & 65.67 & 55.56 & 77.09 & 60.52 & 28M & 4.5G \\
MERSiamC3D\cite{MERSiamC3D} '21 & 81.89 & 83.00 & 87.63 & 88.18 & 68.75 & 64.00 & 72.80 & 74.75 & - & -\\
AU-GCN\cite{AUGCN} '21 & 74.27 & 70.47 & 87.10 & 87.98 & 74.26 & 70.45 & 78.90 & 77.51 & - & -\\
ConvNeXt-T\cite{convnext} '22 & 84.62 & 82.11 & 91.56 & 89.67 & 73.1 & 60.43 & 82.44 & 66.86 & 29M & 4.5G \\
MMNet\cite{li2022mmnet} '22& 86.09 & 85.21 & 92.25 & 93.68 & 80.14 & 72.91 & \underline{90.22} & 83.91 & 10.2M & 7.6G\\
AMAN\cite{AMAN} '22 & 75.40 & 71.25 & - & - & 68.85 & 66.82 & - & - & - & - \\
FRL-DGT\cite{cvpr2023frl} '23 & 75.70 & 74.80 & 88.50 & 89.80 & - & - & - & - & - & - \\
C3DBed\cite{C3DBed} '23 & 77.64 & 75.20 & 88.82 & 89.78 & 75.73 & 72.16 & 80.67 & 81.26 & - & - \\
VMamba-T\cite{vmamba} '23 & 74.49 & 68.08 & 85.71 & 81.55 & 62.69 & 55.40 & 74.05 & 60.56 & 30M & 4.9G \\
$\mu$-BERT†\cite{u-BERT} '23 & †83.48 & \underline{†85.53} & †89.14 & †90.34 & \textbf{†84.75} & \textbf{†83.86} & - & - & - & - \\
ATM-GCN\cite{ATM-GCN} '24& - & - & 90.42 & 90.48 & - & - & 79.20 & 80.49 & - & - \\
DecFlow\cite{DecFlow} '24 & 82.10 & 81.80 & 94.20 & 93.10 & - & - & - & - & - & - \\
SRMCL\cite{SRMCL} '24 & 83.20 & 82.86 & - & - & 74.63& 65.99& 88.66& 84.70& 33.7M & 6.6G\\
TleMer†\cite{TleMer} '24 & \underline{†86.35} & †84.08 & \underline{†94.67} & \underline{†93.40} & \underline{†84.44} & \underline{†81.78} & †90.15 & †84.85 & - & - \\
LTR3O\cite{ltr3o} '25 &  81.78 & 79.05 & - & - & 80.15 & 75.74 & - & - & - & - \\
CSARNet\cite{carnet} '25 &  78.46 & 76.65  & 92.98 & 92.54 & 77.94 & 69.80 & 79.24 & 78.94 & - & - \\
MOL\cite{MOL} '25 & 79.23 & 75.85 & 91.26 & 88.91 & 76.68 & 71.90 &  88.36 & 82.72 & - & - \\
\midrule
GAMDSS(rise) & \textbf{87.50} & \textbf{86.17} & \textbf{94.84} & \textbf{94.30} & 80.15 & 73.35 & \textbf{90.35} & \underline{85.16} & 12.6M & 5.5G\\
GAMDSS(fall) & 76.92 & 74.44 & 92.91 & 92.25 & 75.37 & 67.99 & 84.73 & 76.62 & 12.6M & 5.5G\\
GAMDSS(full) & 87.04 & 85.48 & 94.19 & 93.62 & \textbf{82.84} & \textbf{81.47} & 90.07 & \textbf{85.23} & 12.6M & 11.1G\\
\bottomrule
\multicolumn{8}{l}{† Additional pre-training phase required.}
\end{tabular}
\label{tab:comparison-samm and cas2}
\end{table*}

\begin{table}[t]
\caption{Performance comparison on CAS(ME)$^3$}
\centering
\begin{tabular}{@{}c|cc|cc}
\toprule
\multirow{2}{*}{Model} & \multicolumn{2}{c|}{7-class (\% ↑)} & \multicolumn{2}{c}{4-class (\% ↑)} \\
& UF1 & UAR & UF1 & UAR \\
\midrule
AlexNet\cite{casme3} '22& 17.59 & 18.01 & 29.15 & 29.10 \\
$\mu$-BERT\cite{u-BERT} '23& 32.64 & 32.54 & 47.18 & 49.13 \\
SFAMNet\cite{SFAMNet} '24& 23.65 & 23.73 & 44.62 & 47.97 \\
ATM-GCN\cite{ATM-GCN} '24& 43.08 & 42.83 & 54.23 & 53.49 \\
PC-GCN\cite{PC-GCN} '25& 35.64 & 41.59 & 47.64 & 53.66 \\
\midrule
GAMDSS(rise) & \textbf{53.29} & \textbf{62.73} & \textbf{70.75} & \textbf{76.03}\\
GAMDSS(fall) & 43.33 & 58.27 & 59.09 & 70.47 \\
GAMDSS(full) & 42.58 & 58.72 & 59.82 & 67.18 \\
\bottomrule
\end{tabular}
\label{tab:comparison-cas3}
\end{table}

In accordance with the majority of the studies, the leave-one-subject-out (LOSO) cross-validation method is employed. The accuracy (ACC), unweighted F1 score (UF1) and unweighted average recall (UAR) are used as the evaluation metrics:
\begin{equation}
\text{UF1} = \frac{1}{C} \sum_{i=1}^{C} \frac{2 \times \text{TP}_i}{\text{TP}_i + \text{FP}_i + \text{FN}_i}
\end{equation}
\begin{equation}
\text{UAR} = \frac{1}{C} \sum_{i=1}^{C} \frac{\text{TP}_i}{N_i}
\end{equation}

\subsection{Implementation Details}

In the present experiments, the MTCNN algorithm is utilized for the purpose of face cropping, with all video frames resized to 224 × 224 for the purposes of training and testing. The AdamW optimiser is employed during the training phase in order to optimise the model. The initial learning rate is set to 0.0008, the batch size is 16, the cross-entropy loss function is used as the loss function, and 200 epochs are iterated in a loop for each sample. The model architecture comprises RMT as the temporal branch backbone and ViT as the spatial branch backbone to form the spatio-temporal unit. Furthermore, the selection range factor $\lambda_R$ of the GAMDSS is set to 0.1. All experiments are performed on a single NVIDIA RTX 4060 Ti GPU using the PyTorch framework.

\begin{table*}[!t]
\caption{Ablation study on CASME \uppercase\expandafter{\romannumeral2} and SAMM.}
\centering
\begin{tabular}{c|cc|cc|cc}
\toprule
\multirow{2}{*}{DataSet}& \multicolumn{2}{c|}{Model}& \multicolumn{2}{c|}{7 or 5-class (\% ↑)}&\multicolumn{2}{c}{4 or 3-class (\% ↑)} \\ 
 & D & S & ACC./UAR & UF1 & ACC./UAR & UF1 \\
\midrule
\multirow{4}{*}{CASME \uppercase\expandafter{\romannumeral2}}& $\times$ & $\times$ & 82.19 & 79.80  & 90.90 & 89.69 \\
& $\checkmark$ & $\times$ &85.02(+2.83) & 83.08(+3.28)  & 93.50(+2.60) & 91.81(+2.12) \\
& $\times$ & $\checkmark$ & 85.82(+3.63) & 83.73(+3.93)  & 93.50(+2.60) & 92.25(+2.56) \\
& $\checkmark$ & $\checkmark$ & \textbf{87.04(+4.85)} & \textbf{85.48(+5.68)} & \textbf{94.19(+3.29)} & \textbf{93.62(+3.93)} \\
\midrule
\multirow{4}{*}{SAMM}& $\times$ & $\times$ & 78.35 & 72.56 & 85.49 & 76.76 \\
& $\checkmark$ & $\times$ & 79.85(+1.5) & 76.97(+4.41) & 87.79(+2.3) & 84.64(+7.88) \\
& $\times$ & $\checkmark$ & 79.85(+1.5) & 75.81(+3.25) & 86.25(+0.76) & 79.01(+2.25) \\
& $\checkmark$ & $\checkmark$ & \textbf{82.84(+4.49)} & \textbf{81.47(+8.91)} & \textbf{90.07(+4.58)} & \textbf{85.23(+8.47)} \\
\midrule
\multirow{4}{*}{CAS(ME)$^3$}& $\times$ & $\times$ & 51.69 & 35.54 & 58.84 & 50.58 \\
& $\checkmark$ & $\times$ & 56.02(+4.33) & 38.06(+2.52) & 62.19(+3.35) & 53.54(+2.96) \\
& $\times$ & $\checkmark$ & 57.28(+5.59) & 39.09(+3.55) & 64.21(+5.37) & 55.66(+5.08) \\
& $\checkmark$ & $\checkmark$ & \textbf{58.72(+7.03)} & \textbf{42.58(+7.04)} & \textbf{67.18(+8.34)} & \textbf{59.82(+9.24)} \\
\midrule
\multicolumn{7}{l}{D and S represent the dynamic frame reselection mechanism and the spatial branch.}
\end{tabular}
\label{tab:ablation}
\end{table*}

\begin{table*}[!t]
\caption{Strategies for range of frame reselection.}
\centering
\begin{tabular}{cc|cc|cc|cc|cc}
\toprule
\multicolumn{2}{c|}{$\lambda_{r}$} & \multicolumn{2}{c}{SAMM (\% ↑)} & \multicolumn{2}{c}{CASME \uppercase\expandafter{\romannumeral2}(\% ↑)} & \multicolumn{2}{c}{4DME (\% ↑)} & \multicolumn{2}{c}{MMEW (\% ↑)}\\ 
$\lambda_{rise}$ & $\lambda_{fall}$ & ACC. & UF1 & ACC. & UF1 & ACC. & UF1 & ACC. & UF1 \\
\midrule
0.00 & 0.00 & 79.85	& 75.81 & 85.82 & 83.73 & 85.09 & 68.29 & 79.91 & 64.41\\
0.05 & 0.05 & 81.34 & 78.29 & \textbf{87.04} & \textbf{85.48} & 85.09 & 68.29 & \textbf{80.34} & \textbf{68.20}\\
0.10 & 0.10 & \textbf{82.84} & \textbf{81.47} & 85.02 & 84.44 & \textbf{86.84} & 74.98 & 79.48 & 66.14\\
0.15 & 0.15 & 82.09 & 77.25 & 85.82 & 84.66 & 85.53 & \textbf{77.61} & 79.48 & 62.84\\
0.20 & 0.20 & 81.34 & 77.65 & 85.43 & 82.96 & 82.89 & 69.56 & 79.05 & 61.36\\
0.25 & 0.25 & 78.36 & 74.33 & 82.17 & 79.92 & 83.33 & 69.10 & 78.21 & 55.02\\
\midrule
\end{tabular}
\label{tab:strategy}
\end{table*}

\begin{table*}[!t]
\caption{Backbone study on SAMM and CASME \uppercase\expandafter{\romannumeral2}.}
\centering
\begin{tabular}{@{}c|cc|cc|c|cc|cc|c@{}}
\toprule
\multirow{3}{*}{Model} & \multicolumn{4}{c|}{SAMM} & \multirow{3}{*}{Train Time} & \multicolumn{4}{c|}{CASME \uppercase\expandafter{\romannumeral2}} & \multirow{3}{*}{Train Time}\\
& \multicolumn{2}{c|}{5-class (\% ↑)} & \multicolumn{2}{c|}{3-class (\% ↑)} & &\multicolumn{2}{c|}{5-class (\% ↑)} & \multicolumn{2}{c|}{3-class (\% ↑)} & \\
 & ACC. & UF1 & ACC. & UF1 & & ACC & UF1 & ACC. & UF1 & \\
\midrule
ResNet\cite{resnet} & 75.37 & 66.32 & 83.96 & 74.30& 1 h 10 min 38.97 s & 81.37 & 78.12 & 89.61 & 86.94 & 30 min 7.36 s\\
ResNet (+3DF-N\cite{li2021tip}) & 77.61 & 67.51 & 85.49 & \textbf{79.29} & 2 h 12 min 37.38 s & 80.97 & 77.82 & 88.96 & 86.03 & 1 h 12 min 55.08 s\\
ResNet (+GAMDSS) & \textbf{77.61} & \textbf{69.95} & \textbf{86.25} & 78.50 & 1 h 51 min 54.53 s & \textbf{81.78} & \textbf{79.23} & \textbf{90.90} & \textbf{87.51} & 54 min 13.24 s \\
\midrule
ConvNeXt\cite{convnext} & 73.13 & 60.43 & 82.44 & 66.86 & 1 h 38 min 43.76 s & 84.62 & 82.11 & 91.56 & 89.67 & 46 min 16.48 s \\
ConvNeXt (+3DF-N\cite{li2021tip}) & 72.38 & 62.78 & \textbf{84.73} & \textbf{73.95} & 2 h 40 min 42.17 s & 82.99 & 81.57 & 92.85 & 90.97 & 1 h 29 min 4.20 s \\
ConvNeXt (+GAMDSS) & \textbf{75.37} & \textbf{64.59} & 84.73 & 70.38 & 2 h 31 min 3.35 s & \textbf{85.02} & \textbf{83.85} & \textbf{92.85} & \textbf{91.82} & 1 h 20 min 58.83 s \\
\midrule
Swin-T\cite{swin} & 65.67 & 55.56 & 77.09 & 60.52 & 48 min 42.82 s & 67.21 & 62.47 & 84.42 & 79.87 & 32 min 9.06 s \\
Swin-T (+3DF-N\cite{li2021tip}) & 65.67 & 54.69 & 77.86 & 61.88 & 1 h 50 min 41.22 s & 66.80 & 61.73 & 83.77 & 78.54 & 1 h 14 min 56.78 s \\
Swin-T (+GAMDSS)) & \textbf{67.91} & \textbf{57.78} & \textbf{78.63} & \textbf{62.33} & 1 h 22 min 48.79 s & \textbf{68.02} & \textbf{63.51} & \textbf{85.71} & \textbf{82.13} & 57 min 13.72 s \\
\midrule
VMamba\cite{vmamba} & 62.69 & 55.40 & 74.04 & 57.29 & 1 h 15 min 2.06 s & 74.49 & 68.08 & 85.71 & 81.55 & 40 min 5.61 s \\
VMamba (+3DF-N\cite{li2021tip}) & \textbf{65.67} & 54.69 & 75.57 & 59.43 & 2 h 17 min 0.46 s & 73.68 & 66.74 & 84.42 & 81.03 & 1 h 22 min 53.34 s \\
VMamba (+GAMDSS) & 64.93 & \textbf{58.52} & \textbf{77.09} & \textbf{61.72} & 1 h 58 min 3.19 s & \textbf{75.71} & \textbf{70.97} & \textbf{87.01} & \textbf{83.77} & 1 h 3 min 12.16 s\\
\midrule
RMT\cite{RMT2024} & 79.85 & 75.81 & 86.25 & 79.01 & 2 h 33 min 4.25 s & 85.82 & 83.73 & 93.5 & 92.25 & 1 h 47 min 47.31 s \\
RMT (+3DF-N\cite{li2021tip}) & 81.34 & 76.32 & 87.02 & 81.36 & 3 h 35 min 2.66 s & 85.82 & 83.03 & 92.86 & 91.44 & 2 h 30 min 35.04 s\\
RMT (+GAMDSS) & \textbf{82.84} & \textbf{81.47} & \textbf{90.07} & \textbf{85.23} & 3 h 54 min 11.90 s & \textbf{87.04} & \textbf{85.48} & \textbf{94.19} & \textbf{93.62} & 2 h 41 min 40.97 s\\
\midrule
\end{tabular}
\label{tab:backbone}
\end{table*}

\begin{table*}[!t]
\caption{Feature processing study on the SAMM and CASME \uppercase\expandafter{\romannumeral2}.}
\centering
\begin{tabular}{@{}c|c|cc|cc|c|cc|cc|c}
\toprule
\multirow{3}{*}{Model} &\multirow{3}{*}{Feature} & \multicolumn{4}{c|}{SAMM} & \multirow{3}{*}{Training Time} & \multicolumn{4}{c|}{CASME \uppercase\expandafter{\romannumeral2}} & \multirow{3}{*}{Training Time}\\
& & \multicolumn{2}{c|}{5-class (\% ↑)} & \multicolumn{2}{c|}{3-class (\% ↑)} &  & \multicolumn{2}{c|}{5-class (\% ↑)} & \multicolumn{2}{c|}{3-class (\% ↑)} & \\
& & ACC. & UF1 & ACC. & UF1 &  & ACC & UF1 & ACC. & UF1 & \\
\midrule
\multirow{2}{*}{ResNet\cite{resnet}} & flow & 76.12 & 65.47 & 81.82 & 73.57 & 1 h 44 min 10.57 s & 78.95 & 74.71 & 87.66 & 84.27 & 52 min 31.96 s \\
 & difference & 75.37 & 66.32 & 83.96 & 74.30 & 1 h 8 min 21.87 s & 81.37 & 78.12 & 89.61 & 86.94 & 30 min 7.36 s \\
\midrule
\multirow{2}{*}{ConvNeXt\cite{convnext}} & flow & 62.69 & 52.33 & 77.86 & 61.54 & 2 h 18 min 13.26 s & 63.16 & 53.94 & 72.73 & 60.62 & 1 h 16 min 43.40 s \\
 & difference & 73.13 & 60.43 & 82.44 & 66.86 & 1 h 38 min 43.76 s & 84.62 & 82.11 & 91.56 & 89.67 & 46 min 16.48 s \\
\midrule
\multirow{2}{*}{Swin-T\cite{swin}} & flow & 61.19 & 54.27 & 78.62 & 64.54 & 1 h 2 min 58.61 s & 73.28 & 67.26 & 85.71 & 81.79 & 56 min 22.02 s \\
 & difference & 65.67 & 55.56 & 77.09& 60.52 & 48 min 42.82 s & 67.21 & 62.47 & 84.42 & 79.87 & 32 min 9.06 s \\
\midrule
\multirow{2}{*}{VMamba\cite{vmamba}} & flow & 63.43 & 55.19 & 75.57 & 55.26 & 1 h 41 min 38.12 s & 68.02 & 61.01 & 81.81 & 76.67 & 1 h 13 min 5.91 s \\
 & difference & 62.69 & 55.40 & 74.04 & 57.29 & 1 h 15 min 2.06 s & 74.49 & 68.08 & 85.71 & 81.55 & 40 min 5.61 s \\
\midrule
\multirow{2}{*}{RMT\cite{RMT2024}} & flow & 68.66 & 58.69 & 80.92 & 69.44 & 3 h 25 min 6.90 s & 74.49 & 70.88 & 85.06 & 80.17 & 2 h 31 min 29.94 s \\
 & difference & 79.85 & 75.81 & 86.25 & 79.01 & 2 h 33 min 4.25 s & 85.82 & 83.73 & 93.50 & 92.25 & 1 h 47 min 47.31 s \\
\midrule
\end{tabular}
\label{tab:Feature}
\end{table*}

\subsection{Comparison with the State-Of-The-Art}

Two variants of GAMDSS, designated GAMDSS (rise) and GAMDSS (fall), are developed to investigate the impact of varying stages of micro-expression movement alterations on recognition performance. Tables \ref{tab:comparison-samm and cas2} and \ref{tab:comparison-cas3} present the effectiveness of GAMDSS and its two variants on diverse datasets, offering a comparative analysis with other state-of-the-art methods from the past five years. The results demonstrate that GAMDSS (rise) outperforms GAMDSS (fall) and GAMDSS (full) on the CAS(ME)$^3$ and CASME \uppercase\expandafter{\romannumeral2} datasets, while on the SAMM dataset GAMDSS (full) significantly outperforms GAMDSS (rise) and GAMDSS (fall). Concurrently, Table \ref{tab:comparison-cas3} provides a comprehensive overview of the efficacy of GAMDSS and its two variants on the CAS(ME)$^3$ dataset. Our GAMDSS demonstrates notable proficiency in both the 7-classification and 4-classification tasks, exhibiting substantial superiority over competing methods and attaining optimal outcomes. Specifically, in the 7-classification task, the GAMDSS model enhances the UF1 and UAR metrics by 10.21 and 19.9 percentage points, respectively, in comparison to the second-ranked ATM-GCN method. This superiority is also substantial in the 4-classification task, with enhancements of 16.52\% and 22.54\%, respectively. The model is also tested on smaller datasets, such as CASME \uppercase\expandafter{\romannumeral2} and SAMM, and the results are shown in Table \ref{tab:comparison-samm and cas2}. In the 5-classification task on the CASME \uppercase\expandafter{\romannumeral2} dataset, GAMDSS achieves the best performance, with an improvement of 1.15 percentage points on the ACC metric and 2.09 percentage points on the UF1 metric, when compared to the second-ranked TleMer method. The TleMer method demonstrates the optimal performance, with enhancements of 0.17 and 0.9 percentage points in the three classification tasks, respectively. Although GAMDSS does not demonstrate superior performance in comparison to $\mu$-BERT and TleMer on the SAMM dataset, it is important to acknowledge that both $\mu$-BERT and TleMer necessitate an additional pre-training phase. The results obtained by MMNet are replicated on an NVIDIA RTX 4060 Ti GPU, in accordance with the steps outlined by the original author.

\begin{figure*}[!ht]
  \centering
  \includegraphics[width=0.7\linewidth]{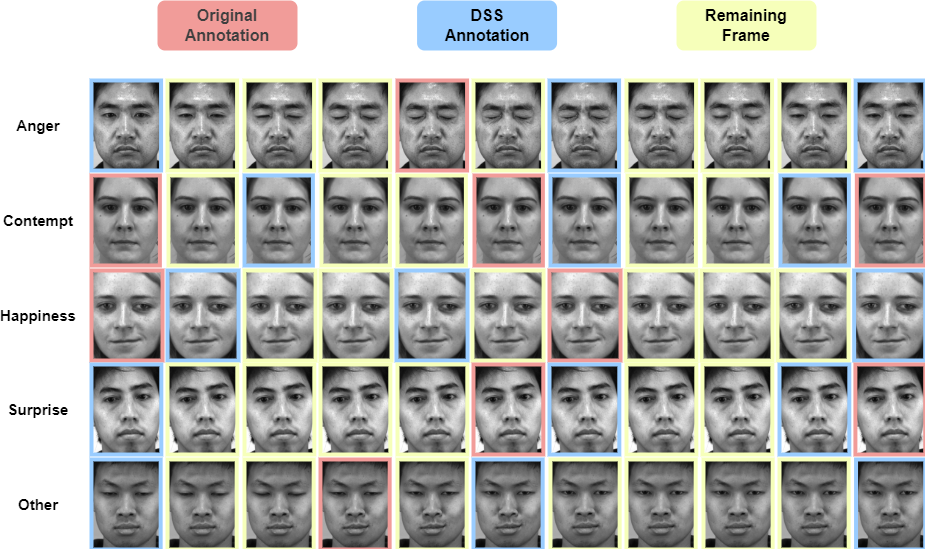}
  \caption{The GAMDSS selection sample is represented visually as follows: pink indicates the original annotation information, blue indicates the GAMDSS re-selected annotation information, and yellow indicates the remaining frames.}
  \label{fig:visualization}
\end{figure*}

\begin{figure}[!t]
  \centering
  \includegraphics[width=0.8\linewidth]{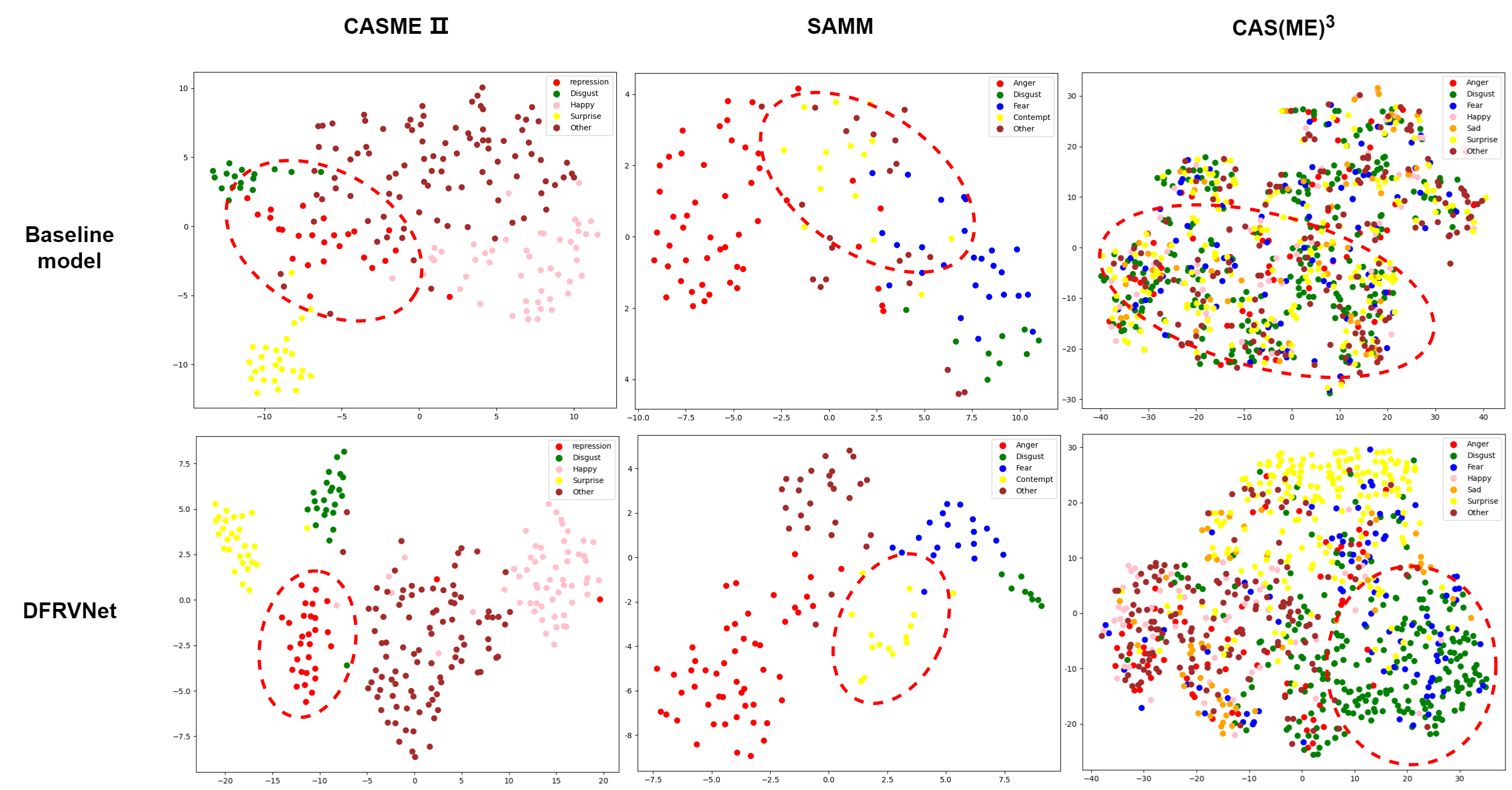}
  \caption{Visualization of feature distributions for the benchmark model and GAMDSS on three datasets. The dashed circles highlight regions where class clusters show clearer separation under GAMDSS compared with the baseline model.}
  \label{fig:tsne}
\end{figure}

\begin{figure*}[!htbp]
    \centering
    \includegraphics[width=0.8\linewidth]{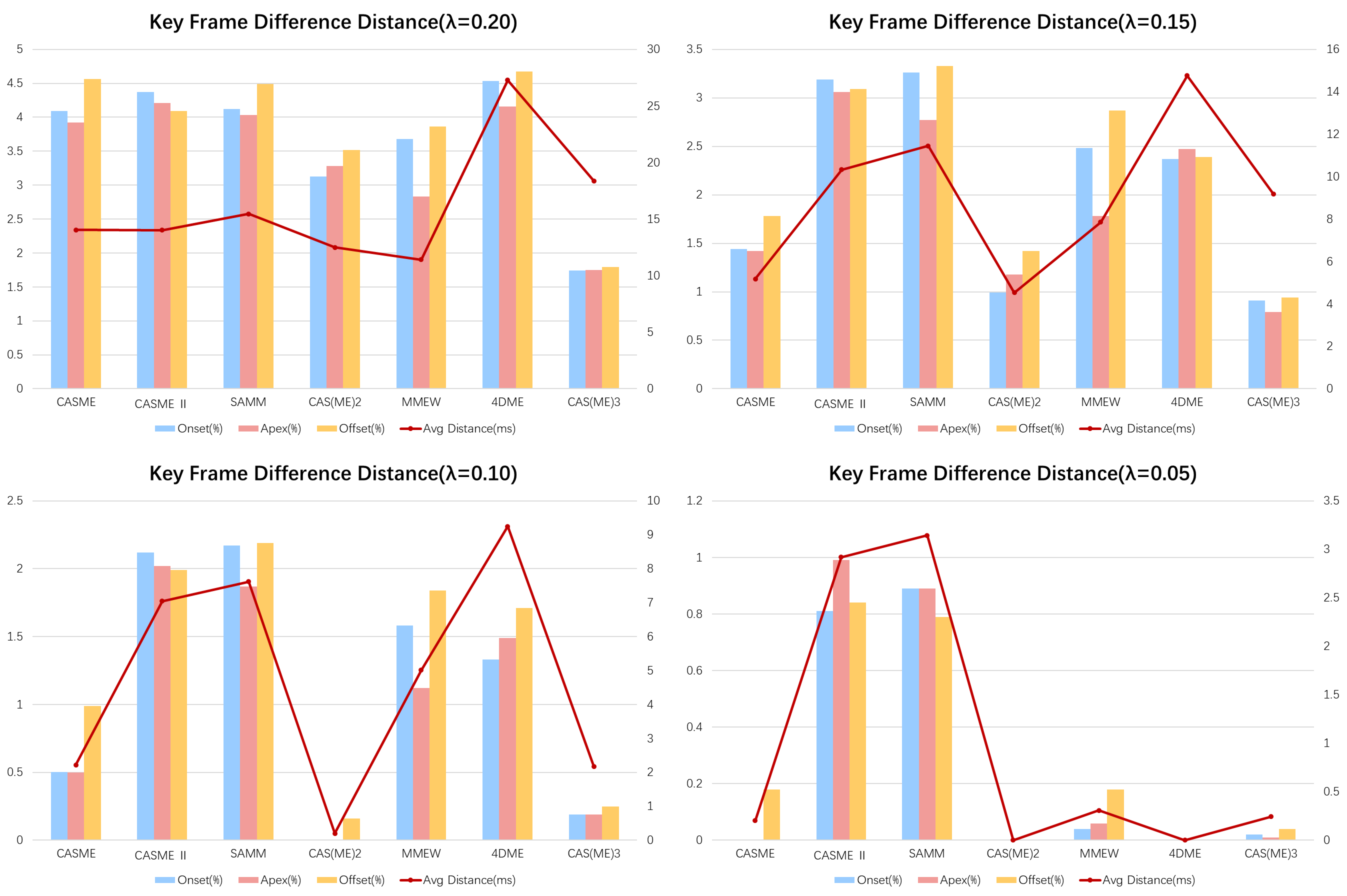}
\caption{Quantitative evaluation of keyframe annotation deviations across datasets. (a) Bar plots show relative deviations ($D\%$) computed via Eq. \ref{d1}. (b) Line plots show the average absolute deviation calculated based on Eq. \ref{d3}, which represents the mean of the absolute deviations for the three key frames at onset, apex, and offset.}
\label{fig:dataset-differ}
\end{figure*}

\begin{figure}[!tbp]
    \centering
    \includegraphics[width=0.7\linewidth]{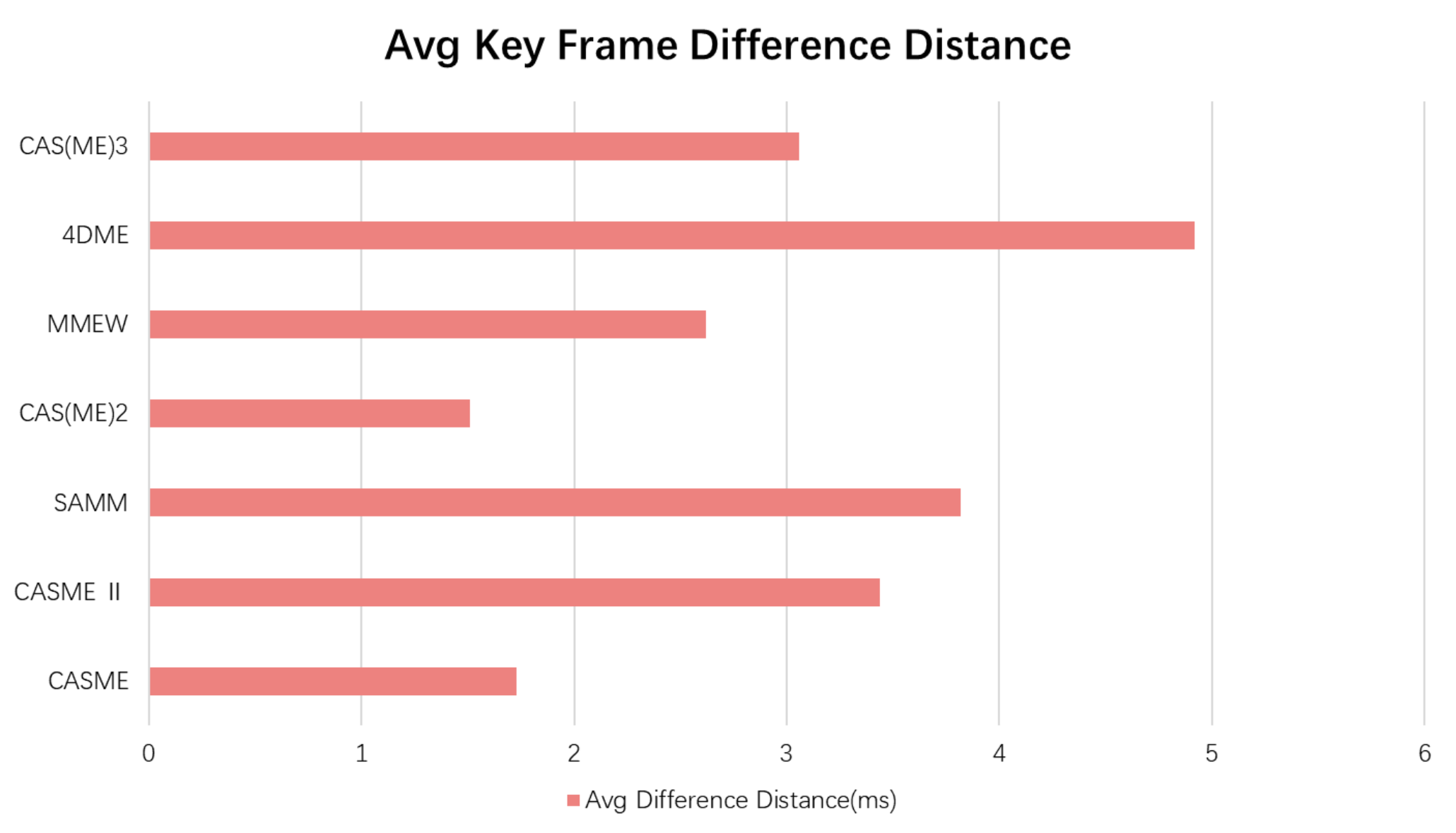}
\caption{Average difference distance between original manual keyframe annotations and GAMDSS re-annotations, in ms.}
\label{fig:differ_change}
\end{figure}

 \subsection{Ablation Study}

\textbf{Network modules.} In order to validate the effectiveness of individual components, four variants are designed: baseline model (W/O D and S), GAMDSS (W/O D), GAMDSS (W/O S), and GAMDSS (complete). The capital letters D and S represent the dynamic frame reselection mechanism and the spatial branch in the S-T unit. Ablation experiments are performed on the CASME \uppercase\expandafter{\romannumeral2}, SAMM and CAS(ME)$^3$ datasets, with the experimental results shown in Table \ref{tab:ablation}. The experimental results demonstrate that incorporating spatial branching within temporal-spatial units leads to a steady enhancement in performance. On the CASME \uppercase\expandafter{\romannumeral2} and SAMM datasets, the addition of spatial branching improves the ACC and UF1 for the 5-classification task by 3.63\%/3.93\% and 1.5\%/3.25\%, respectively. while on the CAS(ME)$^3$ dataset, the UAR and UF1 for the 7-classification task improve by 5.59\%/3.55\%, respectively.

\textbf{Selection of $\lambda$ range.} The results in Table \ref{tab:comparison-samm and cas2} demonstrate that GAMDSS performs differently across datasets with diverse cultural backgrounds. This prompts further investigation into the impact of different R values in GAMDSS on recognition performance. To this end, experiments are conducted on four datasets (two single-cultural background datasets and two multicultural background datasets), with results shown in Table \ref{tab:strategy}. Specifically, in datasets with a multicultural background, such as SAMM and 4DME, larger R values yield better performance. For instance, the performance of SAMM attains its optimal level at R=0.10, while 4DME attains its optimal level at R=0.15. Conversely, in single-cultural context datasets such as CASME \uppercase\expandafter{\romannumeral2} and MMEW, smaller R values prove more effective, with CASME \uppercase\expandafter{\romannumeral2} and MMEW attaining optimal performance at R=0.05. Furthermore, increasing R value beyond its optimal value results in a decline in the model's recognition performance. This decline is attributed to the presence of irrelevant frames with large motion amplitudes in the samples, such as blinking actions. Consequently, GAMDSS erroneously selects this motion information, resulting in performance degradation.

\textbf{Impact of different backbones.} 
To further validate the effectiveness of the GAMDSS method, it is integrated into several mainstream baseline models and compared with the 3DF-N method. The relevant results are shown in Table \ref{tab:backbone}. Specifically, on convolutional neural network models, GAMDSS delivers stable performance improvements. For instance, on the SAMM dataset's five-class classification task, ResNet (+GAMDSS) achieves 2.24\%/3.63\% improvements in ACC./UF1 compared to the original ResNet, and outperforms ResNet (+3DF-N) on the UF1 metric, demonstrating GAMDSS's superiority in key frame selection. Beyond CNNs, we also compared GAMDSS on sequence modeling frameworks like Swin Transformer and VMamba. On the CASME \uppercase\expandafter{\romannumeral2} five-class task, Swin-T (+GAMDSS) outperforms Swin-T (+3DF-N) in ACC./UF1, and similarly achieves superior results on the SAMM dataset. These results indicate that GAMDSS effectively enhances models' ability to extract spatio-temporal dynamic features while mitigating annotation distortions caused by human subjective biases. In particular, GAMDSS achieves performance improvements without introducing additional parameters and integrates seamlessly into any model with just a few lines of code.

\textbf{Feature processing method.} As shown in Table \ref{tab:Feature}, different feature processing methods significantly impact the recognition performance of various mainstream models on both the SAMM and CASME \uppercase\expandafter{\romannumeral2} datasets. Overall, the difference frame method outperforms the optical flow method across most models, achieves higher ACC and UF1 for both 5-class and 3-class tasks while significantly reducing training time. This trend is particularly obvious in models based on convolutional structures. After using difference frames, models not only achieve stable improvements in UF1, such as ResNet increasing from 74.71\% with optical flow to over 78.12\% in the five-classification task on CASME \uppercase\expandafter{\romannumeral2}, but also reduce training time by approximately one-third. This indicates that the local change information provided by differential frames is more direct and less noisy, which enables convolutional models to more effectively capture the subtle dynamics of micro-expressions. Similar patterns are observed in sequence modeling frameworks, indicating that differential frames are applicable not only to traditional convolutional architectures but also to sequence modeling architectures. Notably, RMT inherently preserves temporal dependencies, and difference frames further amplify this advantage, enabling the model to focus more closely on the regions where micro-expressions occur.

\subsection{Visualization}
In order to facilitate a more intuitive analysis of GAMDSS's choice of frames, a selection of samples is drawn from the SAMM dataset to illustrate the discrepancy between the original annotations and those annotated by GAMDSS, see Fig. \ref{fig:visualization}. Within the Anger category, the GAMDSS annotation of Apex frames exhibits a more pronounced squeezing action between the eyebrows compared to the original annotation. Meanwhile, in the Contempt category, which contains a smaller number of samples, the GAMDSS's Onset and Offset frames are less expressive than the original labeling, with no discernible action information. To further analyze the effect of GAMDSS on the features extracted from the model, we employ the t-SNE technique to visualize the distribution of facial expression features extracted from the model. We conduct visualization experiments on three datasets: CAS(ME)$^3$, CASME \uppercase\expandafter{\romannumeral2} and SAMM. The results of the three comparisons are illustrated in Fig. \ref{fig:tsne}. The first row shows the feature distribution of the benchmark model and the second row shows the feature distribution of GAMDSS. It can be seen that for the benchmark model, although emotions can be categorized, the decision boundary is still unclear for similar emotions, such as fear and disgust, showing a certain degree of overlap. The incorporation of GAMDSS serves to refine the decision boundary of the model, thus enhancing its capacity to differentiate between discrete categories. This enhancement is exemplified by the discernment of repression in CASME \uppercase\expandafter{\romannumeral2}, contempt in the SAMM data set and disgust in CAS(ME)$^3$.

\section{Keyframe subjective error}
\label{sec:Keyframe subjective error}

The subjective error of manual annotation is a key challenge faced in the field of micro-expression recognition. To quantify this subjective error, we systematically evaluate the temporal-spatial deviation between the original annotation and the re-annotation using GAMDSS on seven mainstream datasets. The experiments dynamically adjust the frame pair search range by adjusting the differential range parameter $\lambda$ (0.2, 0.15, 0.1, 0.05, respectively) and compute the deviation between the re-selected frames and the original manually labeled ones using the original manually labeled Onset frames, the Apex frames, and the Offset frames as a baseline. Two metrics are used for the deviation metric, which are the percentage distance D (\%) relative to the sequence length; and the absolute time displacement D (ms). They can be expressed by the following equations:

\begin{equation}
D(\%) = \frac{\left| f_{\text{re}} - f_{\text{orig}} \right|}{L} \times 100\%
\label{d1}
\end{equation}

\begin{equation}
D(ms) = \left| f_{\text{re}} - f_{\text{orig}} \right| \times \frac{1000}{\text{FPS}}
\label{d2}
\end{equation}

\begin{equation}
\overline{D}_{\text{ms}} = \frac{1}{|\mathcal{K}|} \sum_{k \in \mathcal{K}} D_{\text{ms}}^{(k)}, \quad \mathcal{K} = \{\text{Onset, Apex, Offset}\}
\label{d3}
\end{equation}

Where \( D(\%) \) is the relative deviation computed via Eq. \ref{d1}, \( D(ms) \) is the absolute deviation for a single keyframe computed via Eq. \ref{d2}, \( \overline{D}_{\text{ms}} \) is the average absolute deviation across the three keyframes as defined in Eq. \ref{d3}, \( \mathcal{K} = \{\text{Onset, Apex, Offset}\} \) denotes the set of keyframe types, \( D_{\text{ms}}^{(k)} \) represents the absolute deviation for keyframe type \( k \in \mathcal{K} \), \( |\mathcal{K}| \) is the cardinality of \( \mathcal{K} \) (here \( |\mathcal{K}| = 3 \)), \( f_{re} \) is the index of the GAMDSS re-selection annotation, \( f_{orig} \) is the index of the original annotation, \( L \) is the total number of frames in the micro-expression sequence, and FPS is the frame rate of the dataset acquisition.

The experimental results are shown in Fig. \ref{fig:dataset-differ}. First, under the relaxed search condition of $\lambda$ = 0.20, the D (\%) of all datasets exceeds 3.5\%, and the corresponding D (ms) exceeds 16 ms. This highlights that under relaxed conditions, GAMDSS tends to misclassify irrelevant actions such as blinking and head shaking as key frames. When $\lambda$ is reduced to 0.15, the D (\%) for all datasets decreases to 2\%, with the corresponding D (ms) decreasing to 9ms, representing a reduction of approximately 57\%. This indicates that reducing the search range can effectively suppress interference from irrelevant actions. Further reducing $\lambda$ to 0.10, the D (\%) for single-culture datasets such as CAS(ME)$^2$ and CAS(ME)3 decreases to 0.86\%, with the corresponding D (ms) decreasing to 3.26ms. Meanwhile, the D (\%) for multi-culture datasets such as SAMM and 4DME remains at 1.75\%, with the corresponding D (ms) remaining at 8.4ms. The difference between the two begins to differ. When $\lambda$ is finally reduced to 0.05, the average relative offset for multiple single-culture datasets such as CASME and CAS(ME)$^2$ has decreases to near 0\%, at which point the re-selected frames highly overlap with the original frame annotations. However, for multicultural datasets such as the SAMM dataset, D (ms) remains at 3 ms. This indicates that under strict selection conditions, cultural diversity still influences the annotation of key frames.

Generally, the smaller the distance between GT key frames and manually annotated key frames, the higher the annotation quality of the dataset. At this point, when changes in the local search range of GAMDSS do not significantly alter the gap between re-labeled key frames and manually labeled key frames, we use this as a theoretical basis to analyze the change in the difference distance of key frames as $\lambda$ increases, and convert the results into absolute deviation. Since at $\lambda$=0.20, GAMDSS misclassifies irrelevant actions such as blinking and head shaking as key frames. Therefore, we restrict the key frame difference distribution to a limited range of 0 to 0.15, with the results shown in Fig. \ref{fig:differ_change}. The results indicate that the increase in differences for the multicultural dataset is significant, with an average increase of 4.36 ms. In contrast, the increase in differences for the single-culture dataset is smaller, with an average increase of 2.4 ms. This suggests that the annotation quality of the multicultural dataset is lower than that of the single-culture dataset.

\section{Discussion}
\label{sec:Discussion}

Further comparison of the performance contributions of GAMDSS (W/O D) and GAMDSS (W/O S) reveals that their performance contributions differ on different datasets. This variation can be attributed to cultural differences within the datasets. Specifically, the participants in the CAS(ME)$^3$ and CASME \uppercase\expandafter{\romannumeral2} datasets are all Asian, with an average age between 20 and 25, while the participants in the SAMM dataset come from diverse cultural backgrounds and have a wider age range. The diversity in culture and age may increase the difficulty of labeling in the datasets, resulting in a more pronounced performance improvement of GAMDSS on the SAMM dataset. Furthermore, the results in Table \ref{tab:comparison-samm and cas2} validated our view: on the CAS(ME)$^3$ and CASME \uppercase\expandafter{\romannumeral2} datasets, GAMDSS (rise) outperformed GAMDSS (full), while on the SAMM dataset, the opposite is true, with GAMDSS (full) outperforming GAMDSS (rise). This finding suggests that, under the original annotation conditions, the SAMM dataset may contain more valuable spatio-temporal motion information during the descent phase. This implies that the ground truth Apex frames may be located after the annotated Apex frames. 

This phenomenon can be explained by micro-expression movement patterns, as shown in Fig. \ref{fig:dataset-differ}. In single culture datasets, subjects' facial muscle movement patterns exhibit high homogeneity, with micro-expression intensity trajectories often showing symmetrical trends. Thus, focusing solely on the rise phase suffices to capture key muscle movement evolution information. However, in cross-cultural datasets, differing cultural backgrounds lead to systematic variations in micro-expression intensity trajectories, duration, and even local muscle coordination patterns. The complete “rise-fall” dynamic cycle exhibits a trend of initial suppression followed by amplification. In such cases, modeling and comparing the entire motion cycle better corrects these systematic annotation biases, resulting in improved robustness on datasets like SAMM. Furthermore, although CAS(ME)$^3$ exhibits low relative deviation under different $\lambda$ conditions, subjective errors still exist, which may stem from its complex motion patterns under high ecological induction paradigms.

However, despite proposing the GAMDSS framework and experimentally validating its effectiveness in mitigating subjective errors in keyframe annotation, several limitations warrant discussion. First, our method still relies on original manual annotations. While the micro-expression dataset used in this study typically does not contain severely biased annotations, we acknowledge this limitation. To address potentially severe deviation scenarios, a viable improvement direction is to explore how to integrate semantic information from facial movements, such as action units, to assist the adjustment process, thereby further enhancing the model's robustness. Secondly, another viable direction involves integrating GAMDSS with micro-expression spotting (MES) methods to reduce reliance on manual annotations. Finally, the datasets currently employed are collected under laboratory-induced conditions, whereas in real-world scenarios, micro-expressions often co-occur with macro-expressions. In such contexts, GAMDSS may misclassify macro-expressions as key micro-expression frames. To mitigate this issue, a more optimal approach involves setting threshold windows to restrict selection ranges, thereby improving the model's recognition capabilities in authentic settings.

\section{Conclusion}
\label{sec:Conclusion}
Micro-expressions are unconscious external manifestations of human emotions and hold significant application value in the field of clinical psychology. However, traditional manual annotation is prone to the subjective influence of annotators, resulting in certain annotation biases. To address this issue, we propose a novel global anti-monotonic difference selection strategy framework, GAMDSS. It re-selects the onset and apex frames with the greatest micro-expression differences from complete micro-expression action sequences and determines the offset frame based on this, thereby constructing rich spatio-temporal dynamic features. To demonstrate the advantages of GAMDSS and validate its effectiveness, we conducted extensive experiments on seven widely recognized micro-expression datasets. The experimental results showed that for micro-expression datasets from a single cultural background, such as CASME \uppercase\expandafter{\romannumeral2} and CAS(ME)$^3$, the onset frame and apex frame framework is sufficient to capture most micro-expression change features. However, for cross-cultural micro-expression datasets, such as SAMM and 4DME, this assumption no longer holds, necessitating the use of differential frame resampling to compute the complete motion changes required for micro-expression recognition. These findings directly support our argument for re-examining the effectiveness and universality of the current micro-expression dataset annotation paradigm.
Currently, our method still relies on manual annotation. In the future, we plan to combine this method with the MES method to reduce reliance on manual annotation. Additionally, we aim to integrate this method with macro-expression recognition techniques to enhance micro-expression recognition capabilities in real-world scenarios. These improvements may further enhance micro-expression recognition performance and facilitate the practical application of micro-expression recognition technologies in real-world settings.


\bibliographystyle{IEEEtran}
\bibliography{myreference}




\begin{IEEEbiography}[{\includegraphics[width=1in,height=1.25in,clip,keepaspectratio]{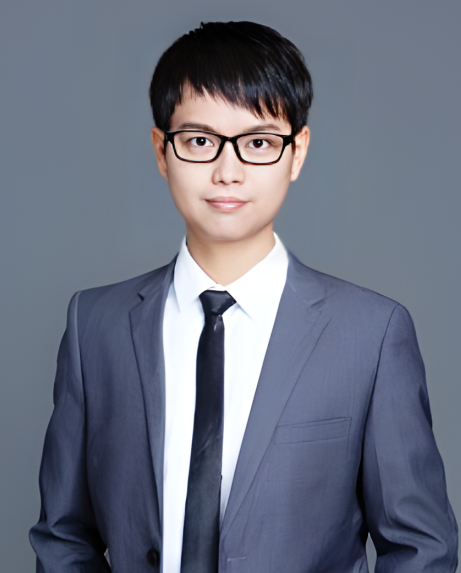}}]{Dr. Feng Liu, IEEE Senior Member} is a research assistant professor at School of Psychology, Shanghai Jiao Tong University. He is an Area Editor of CAAI Artificial Intelligence Research, and also as a senior member of China Computer Federation, a member of the Chinese Psychological Society, a professional member of the Emotional Intelligence Committee of Chinese Association for Artificial Intelligence and the Emotional Computing Committee of Chinese Information Processing Society of China. Also as reviewer for numerous IEEE Trans journals including and not limited to: T-PAMI, T-AFFC, T-IP, T-EC, T-CSVT, T-ITS, T-II, T-AI. His research interests include affective computing, computational psychology, computational psychiatry and computational social science.
\end{IEEEbiography}

\begin{IEEEbiography}[{\includegraphics[width=1in,height=1.25in,clip,keepaspectratio]{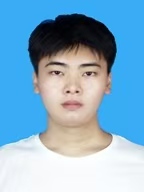}}]{Bingyu Nan} received his B.S. in Software Engineering from Henan University of Science and Technology in 2023. He is currently studying for his M.S. in Software Engineering at Jiangnan University. His research interests include micro-expression recognition, affective computing, and computer vision..
 \end{IEEEbiography}

\begin{IEEEbiography}[{\includegraphics[width=1in,height=1.25in,clip,keepaspectratio]{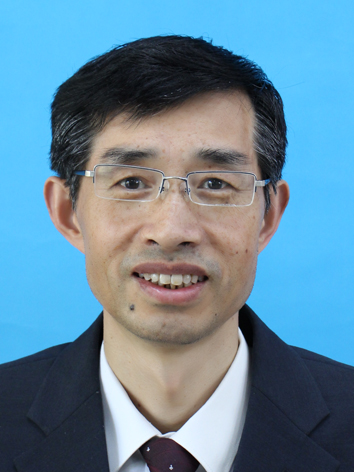}}]{Xuezhong Qian} is an associate professor at the School of Artificial Intelligence and Computer Science at Jiangnan University. He received his master's degree in automation from Jiangnan University in 1998. He is a senior member of the Chinese Computer Society. His research interests include data mining, database technology, and artificial intelligence.
\end{IEEEbiography}

\begin{IEEEbiography}[{\includegraphics[width=1in,height=1.25in,clip,keepaspectratio]{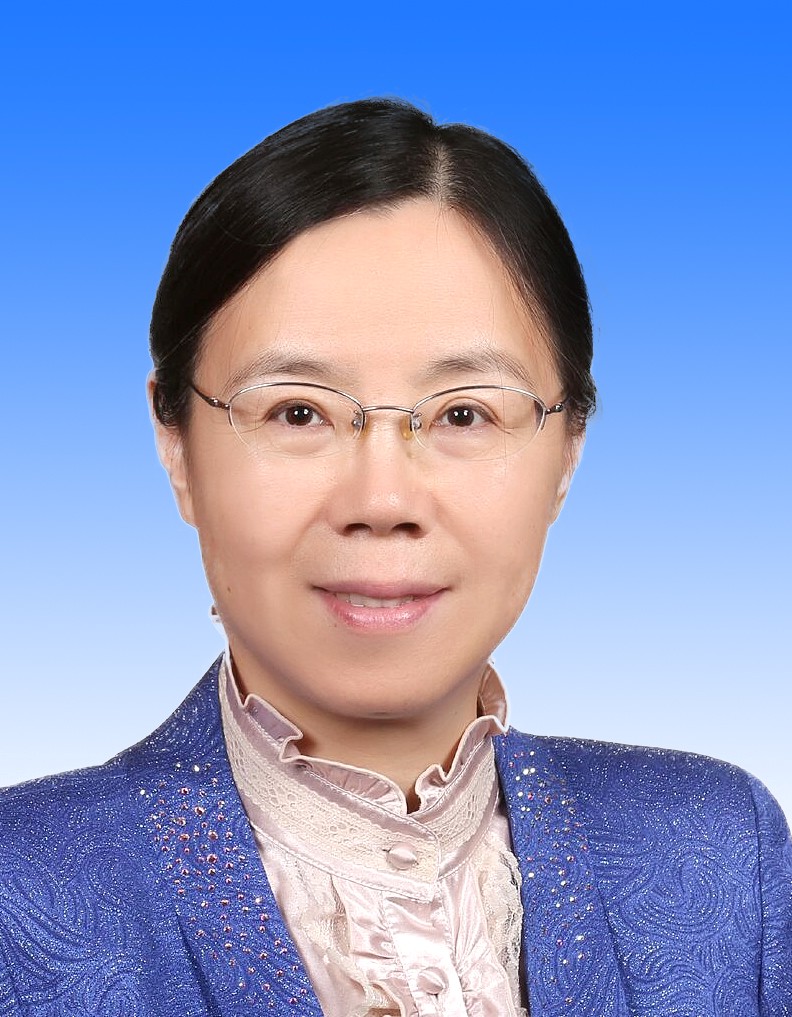}}]{Dr. Xiaolan Fu} received her Ph. D. degree in 1990 from Institute of Psychology, Chinese Academy of Sciences. Currently, she is a Senior Researcher at Cognitive Psychology. Her research interests include visual and computational cognition: (1) attention and perception, (2) learning and memory, and (3) affective computing. At present, she is the dean of School of Psychology, Shanghai Jiao Tong University.
\end{IEEEbiography}


\vfill

\end{document}